\pdfoutput=1

\documentclass[11pt]{article}

\usepackage[]{acl}

\usepackage{times}
\usepackage{latexsym}

\usepackage[T1]{fontenc}

\usepackage[utf8]{inputenc}

\usepackage{microtype}
\usepackage{graphicx}
\usepackage{subfigure}
\usepackage{amsmath}
\usepackage{bbm}
\usepackage{bm}
\usepackage{algorithm}
\usepackage{algorithmic}
\usepackage{amssymb}
\usepackage{array}
\usepackage{booktabs}
\usepackage{multirow}
\usepackage{dutchcal}
\usepackage{hyperref}

\newcommand{\modelname}{\textbf{RSTC}}
\title{Robust Representation Learning with Reliable Pseudo-labels Generation via Self-Adaptive Optimal Transport for Short Text Clustering}

\author{
Xiaolin Zheng, Mengling Hu, Weiming Liu, Chaochao Chen\Thanks{Corresponding author.}, and Xinting Liao\\
Zhejiang University, China\\
\texttt{\{xlzheng,humengling,21831010,zjuccc,xintingliao\}@zju.edu.cn}
}

\begin{document}

\maketitle

\begin{abstract}
Short text clustering is challenging since it takes imbalanced and noisy data as inputs.
Existing approaches cannot solve this problem well, since (1) they are prone to obtain degenerate solutions especially on heavy imbalanced datasets, and (2) they are vulnerable to noises. 
To tackle the above issues, we propose a Robust Short Text Clustering (\modelname) model to improve robustness against imbalanced and noisy data.
\modelname~includes two modules, i.e., \textit{pseudo-label generation module} and \textit{robust representation learning module}.
The former generates pseudo-labels to provide supervision for the later, which contributes to more robust representations and correctly separated clusters.
To provide robustness against the imbalance in data, we propose self-adaptive optimal transport in the pseudo-label generation module.
To improve robustness against the noise in data, we further introduce both class-wise and instance-wise contrastive learning in the robust representation learning module.
Our empirical studies on eight short text clustering datasets demonstrate that \modelname~significantly outperforms the state-of-the-art models.
The code is available at: \href{https://github.com/hmllmh/RSTC}{https://github.com/hmllmh/RSTC}.
\end{abstract}

\section{Introduction}
Text clustering, one of the most fundamental tasks in text mining, aims to group text instances into clusters in an unsupervised manner.
It has been proven to be beneficial in many applications, such as, recommendation system \cite{liu2021leveraging,liu2022collaborative,liu2022exploiting}, opinion mining \cite{stieglitz2018social}, stance detection \cite{li2022unsupervised}, etc.
With the advent of digital era, more and more people enjoy sharing and discovering various of contents on the web, where short text is an import form of information carrier.
Therefore, it is helpful to utilize short text clustering for mining valuable insights on the web.

However, short text clustering is not a trivial task.
On the one hand, short text has many categories and the category distributions are diversifying, where the heavy imbalanced data is common.
The heavy imbalanced data is prone to lead to degenerate solutions where the tail clusters (i.e., the clusters with a small proportion of instances) disappear.
Specifically, the recent deep joint clustering methods for short text clustering, \cite{hadifar2019self} and \cite{zhang2021supporting}, adopt the clustering objective proposed in \cite{xie2016unsupervised}, which may obtain a trivial solution where all the text instances fall into the same cluster \cite{yang2017towards, ji2019invariant}. 
\cite{zhang2021supporting} introduces instance-wise contrastive learning to train discriminative representations, which avoids the trivial solution to some extent.
However, \cite{zhang2021supporting} still tends to generate degenerate solutions, especially on the heavy imbalanced datasets.

On the other hand, short text is typically characterized by noises, which may lead to meaningless or vague representations and thus hurt clustering accuracy and stability.
Existing short text clustering methods cope with the noise problem in three ways, i.e., (1) text preprocessing, (2) outliers postprocessing, and (3) model robustness.
Specifically, earlier methods \cite{xu2017self, hadifar2019self} apply preprocessing procedures on the text \cite{hacohen2020influence} for reducing the negative impact of noises.
The recent method \cite{rakib2020enhancement} proposes to postprocess outliers by repeatedly reassigning outliers to clusters for enhancing the clustering performance.
However, both preprocessing and postprocessing methods do not provide model robustness against the noise in data.
The more recently short text clustering method SCCL \cite{zhang2021supporting} proposes to utilize the instance-wise contrastive learning to support clustering, which is useful for dealing with the noises in the perspective of model robustness.
However, the learned representations of SCCL lack discriminability due to the lack of supervision information, causing insufficiently robust representations.

In summary, there are two main challenges, i.e., 
\textbf{CH1}: How to provide model robustness to the imbalance in data, and avoid the clustering degeneracy? 
\textbf{CH2}: How to improve model robustness against the noise in data, and enhance the clustering performance?

To address the aforementioned issues, in this paper, we propose \modelname, an end-to-end model for short text clustering.
In order to improve model robustness to the imbalance in data (solving \textbf{CH1}) and the noise in data (solving \textbf{CH2}), we utilize two modules in \modelname, i.e., \textit{pseudo-label generation module} and \textit{robust representation learning module}.
The pseudo-label generation module generates pseudo-labels for the original texts.
The robust representation learning module uses the generated pseudo-labels as supervision to facilitate intra-cluster compactness and inter-cluster separability,
thus attaining more robust representations and more correctly separated clusters.
The better cluster predictions in turn can be conductive to generate more reliable pseudo-labels.
The iterative training process forms a virtuous circle, that is, the learned representations and cluster predictions will constantly boost each other, as more reliable pseudo-labels are discovered during iterations.

The key idea to solve \textbf{CH1} is to enforce a constraint on pseudo-labels, i.e., the distribution of the generated pseudo-labels should match the estimated class distribution.
The estimated class distribution is dynamically updated and expected to get closer to the ground truth progressively.
Meanwhile, the estimated class distribution are encouraged to be a uniform distribution for avoiding clustering degeneracy.
We formalize the idea as a new paradigm of optimal transport \cite{peyre2019computational} and the optimization objective can be tractably solved by the Sinkhorn-Knopp \cite{cuturi2013sinkhorn} style algorithm, which needs only a few computational overheads.
For addressing \textbf{CH2}, we further introduce \textit{class-wise} contrastive learning and \textit{instance-wise} contrastive learning in the robust representation learning module.
The class-wise contrastive learning aims to use the pseudo-labels as supervision for achieving smaller intra-cluster distance and larger inter-cluster distance.
While the instance-wise contrastive learning tends to disperse the representations of different instances apart for the separation of overlapped clusters.
These two modules cooperate with each other to provide better short text clustering performance.

We summarize our main contributions as follows: 
(1) We propose an end-to-end model, i.e., \modelname, for short text clustering, the key idea is to discover the pseudo-labels to provide supervision for robust representation learning, hence enhancing the clustering performance.
(2) To our best knowledge, we are the first to propose self-adaptive optimal transport for discovering the pseudo-label information, which provides robustness against the imbalance in data.
(3) We propose the combination of class-wise contrastive learning and instance-wise contrastive learning for robustness against the noise in data.
(4) We conduct extensive experiments on eight short text clustering datasets and the results demonstrate the superiority of \modelname.
\section{Related Work}
\subsection{Short Text Clustering}

Short text clustering is not trivial due to imbalanced  and noisy data.
The existing short text clustering methods can be divided into tree kinds: (1) traditional methods, (2) deep learning methods, and (3) deep joint clustering methods.
The traditional methods \cite{scott1998text, salton1983introduction} often obtain very sparse representations that lack discriminations.
The deep learning method \cite{xu2017self} leverages pre-trained word embeddings \cite{mikolov2013distributed} and deep neural network to enrich the representations.
However, the learned representations may not appropriate for clustering.
The deep joint clustering methods \citet{hadifar2019self, zhang2021supporting} integrate clustering with deep representation learning to learn the representations that are appropriate for clustering.
Moreover, \citet{zhang2021supporting} utilizes the pre-trained SBERT \cite{reimers2019sentence} and contrastive learning to learn discriminative representations, which is conductive to deal with the noises.
However, the adopted clustering objectives are prone to obtain degenerate solutions \cite{yang2017towards, ji2019invariant}, especially on heavy imbalance data.

Among the above methods, only \citet{zhang2021supporting} provides model robustness to the noise in data.
However, its robustness is still insufficient due to the lack of supervision information.
Besides, \citet{zhang2021supporting} cannot deal with various imbalanced data due to the degeneracy problem. 
As a contrast, in this work, we adopt pseudo-label technology to provide reliable supervision to learn robust representations for coping with imbalanced and noisy data.

\subsection{Pseudo-labels for Unsupervised Learning}
Pseudo-labels can be helpful to learn more discriminative representations in unsupervised learning \cite{hu2021learning}.
\citet{caron2018deep} shows that k-means clustering can be utilized to generate pseudo-labels for learning visual representations.
However, it does not have a unified, well-defined objective to optimize (i.e., there are two objectives: k-means loss minimization and cross-entropy loss minimization), which means that it is difficult to characterize its convergence properties.
\citet{asano2020self} proposes SeLa to optimize the same objective (i.e., cross-entropy loss minimization) for both pseudo-label generation and representation learning, which can guarantee its convergence.
Besides, SeLa transforms pseudo-label generation problem into an optimal transport problem.
\citet{caron2020unsupervised} proposes SwAV which combines SeLa with contrastive learning to learn visual representations in an online fashion.
%
%
However, both SeLa and SwAV add the constraint that the distribution of generated pseudo-labels should match the uniform distribution, to avoid clustering degeneracy.
With the constraint, it is hard for them to cope with imbalanced data.
As a contrast, in this work, we propose self-adaptive optimal transport to simultaneously estimate the real class distribution and generate pseudo-labels.
Our method enforce the distribution of the generated pseudo-labels to match the estimated class distribution, and thus can avoid clustering degeneracy and adapt to various imbalanced data.

\section{Methodology}

\begin{figure*}[t]
  \centering
  \includegraphics[width=1\linewidth]{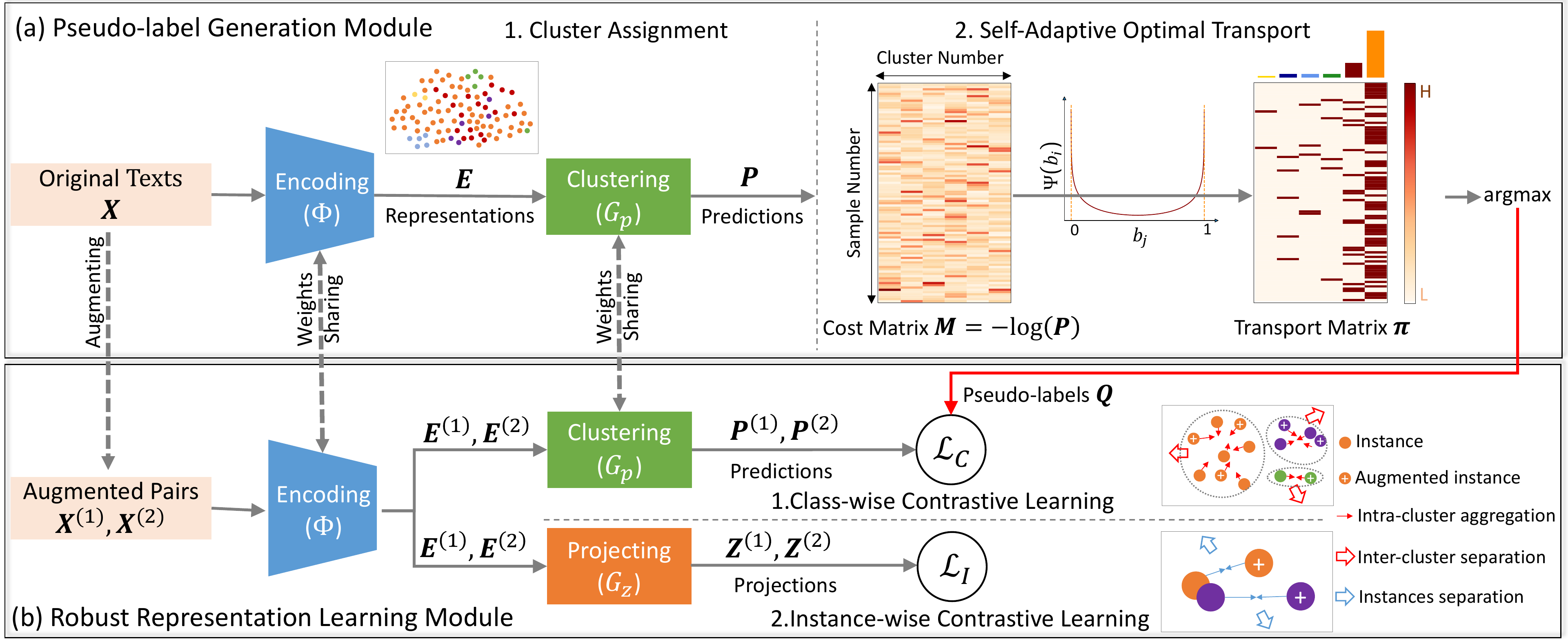}
  \vspace{-0.5cm}
  \caption{{An overview of \modelname, which contains two modules: (a) Pseudo-label Generation Module, and (b) Robust Representation Learning Module.} 
  }
  \vspace{-0.4cm}
  \label{fig:model}
\end{figure*}

\subsection{An Overview of \modelname}
The goal of \modelname~is to discover and utilize the pseudo-labels to provide supervision for robust representation learning.
\modelname~consists of \textit{pseudo-label generation module} and \textit{robust representation learning module}, as illustrated in Fig.\ref{fig:model}.
The pseudo-label generation module aims to generate reliable pseudo-labels for the robust representation learning module.
To achieve this aim, we first obtain cluster predictions by the cluster assignment step, then we excavate pseudo-label information from the predictions by the self-adaptive optimal transport (SAOT) step.
The robust representation learning module aims to use the generated pseudo-labels as supervision to train robust representations.
To achieve this goal, we introduce class-wise and instance-wise contrastive learning.
In this way, \modelname~can provide robustness to imbalanced and noisy data, thus enhancing the clustering performance.

\subsection{Pseudo-label Generation Module}
\label{sec:saot}
We first introduce the pseudo-label generation module.
Although the deep joint clustering methods \cite{xie2016unsupervised, hadifar2019self, zhang2021supporting} are popular these days, their clustering performance is limited due to the following reasons.
Firstly, lacking supervision information prevents the deep joint clustering methods from learning more discriminative representations \cite{hu2021learning}.
Secondly, they are prone to obtain degenerate solutions \cite{yang2017towards,ji2019invariant}, especially on heavy imbalanced datasets.
Therefore, to provide reliable supervision information for various imbalanced data, we propose SAOT in the pseudo-label generation module to generate pseudo-labels for the robust representation learning module.
The overview of pseudo-label generation module is shown in Fig.\ref{fig:model}(a),
which mainly has two steps: \textbf{Step 1:} cluster assignment, and \textbf{Step 2:} SAOT.

\textbf{Step 1:} \textit{cluster assignment.} 
Cluster assignment aims to obtain cluster predictions of the original texts.
Specifically, we adopt SBERT \cite{reimers2019sentence} as the encoding network $\Phi$ to encode the original text $\bm{X}$ as $\Phi(\bm{X})=\bm{E}\in\mathbbm{R}^{N\times D_1}$ where $N$ denotes batch size and $D_1$ is the dimension of the representations. 
We utilize the fully connected layers as the clustering network $G_p$ to predict the cluster assignment probability (predictions), i.e., $G_p(\bm{E})=\bm{P}\in\mathbbm{R}^{N\times C}$, where $C$ is the category number.
The encoding network and the clustering network are fixed in this module.

\textbf{Step 2:} \textit{SAOT.}
SAOT aims to exploit the cluster predictions to discover reliable pseudo-label.
\citet{asano2020self} extends standard cross-entropy minimization to an optimal transport (OT) problem to generate pseudo-labels for learning image representations.
This OT problem can be regarded as seeking the solution of transporting the sample distribution to the class distribution. 
However, the class distribution is unknown.
Although \citet{asano2020self} sets it to a uniform distribution to avoid degenerate solutions, the mismatched class distribution will lead to unreliable pseudo-labels.
Therefore, it is essential to estimate real class distribution for addressing this issue.
The recent research \cite{wang2022solar} studies the class distribution estimation, but it tends to cause clustering degeneracy on heavy imbalanced data, which we will further discuss in Appendix \ref{sec:apph}.
Hence, to discover reliable pseudo-labels on various imbalanced data, we propose SAOT.

We will provide the details of SAOT below.
We expect to minimize the cross entropy loss to generate the pseudo-labels by solving a discrete OT problem.
Specifically, we denote the pseudo-labels as $\bm{Q}\in\mathbbm{R}^{N\times C}$.
Let $\bm{\pi}=\frac{1}{N}\bm{Q}$ be the transport matrix between samples and classes, $\bm{M}=-\log{\bm{P}}$ be the cost matrix to move probability mass from samples to classes. 
The reason that we use $\frac{1}{N}$ between $\bm{\pi}$ and $\bm{Q}$ is the transport matrix should be a joint probability \cite{cuturi2013sinkhorn}, i.e., the sun of all values in the $\bm{\pi}$ should be $1$, while the sum of each raw in $\bm{Q}$ is $1$.
We have, $\bm{Q}^*=\underset{\bm{Q}}{\mathrm{argmin}}\langle \bm{Q}, -\log{\bm{P}}\rangle=N\underset{\bm{\pi}}{\mathrm{argmin}}\langle \bm{\pi}, \bm{M}\rangle$. 
Thus, the OT problem is as follows:
\begin{equation}
    \begin{aligned}
        &\min_{\bm{\pi}}{\langle \bm{\pi}, \bm{M} \rangle} + \epsilon H(\bm{\pi})\\
        &s.t.\,\,\bm{\pi}\bm{1}=\bm{a},\bm{\pi}^T\bm{1}=\bm{b},\bm{\pi}\geq 0,
    \end{aligned}
\end{equation}
where $\epsilon$ is a balance hyper parameter, $H(\bm{\pi})=\langle \bm{\pi},\log\bm{\pi}-1 \rangle$ is the entropy regularization \cite{cuturi2013sinkhorn}, $\bm{a}=\frac{1}{N}\bm{1}$ is the sample distribution, and $\bm{b}$ is an unknown class distribution.
To avoid clustering degeneracy and obtain reliable transport matrix with randomly initialized $\bm{b}$, we introduce a penalty function about $\bm{b}$ to the OT objective and update $\bm{b}$ during the process of solving the transport matrix.
We formulate the SAOT optimization problem as:
\begin{equation}
\label{eq2:saot}
    \begin{aligned}
        &\min_{\bm{\pi},\bm{b}}{\langle \bm{\pi}, \bm{M} \rangle} + \epsilon_1 H(\bm{\pi}) + \epsilon_2 (\Psi(\bm{b}))^T\bm{1}\\
        &s.t.\,\,\bm{\pi}\bm{1}=\bm{a},\bm{\pi}^T\bm{1}=\bm{b}, \bm{\pi}\geq0, \bm{b}^T\bm{1}=1,
    \end{aligned}
\end{equation}
where $\epsilon_1$ and $\epsilon_2$ are balance hyper-parameters, $\Psi(\bm{b})=-\log\bm{b} - \log(1-\bm{b})$ is the penalty function about $\bm{b}$.
The penalty function not only limits $b_j$ (a value of $\bm{b}$) ranges from $0$ to $1$, but also avoids clustering degeneracy by encouraging $\bm{b}$ to be a uniform distribution.
The encouragement is achieved by increasing the punishment for $b_j$ that is close to $0$ or $1$.
Besides, the level of the encouragement can be adjusted by $\epsilon_2$.
Specifically, there are two critical terms in Equation (\ref{eq2:saot}) for exploring $\bm{b}$, i.e., (1) the cost matrix $\bm{M}$ and (2) the penalty function $\Psi(\bm{b})$, and we use $\epsilon_2$ to balance these two terms.
For balanced data, both $\bm{M}$ and $\Psi(\bm{b})$ encourage $\bm{b}$ to be a uniform distribution.
For imbalanced data, $\bm{M}$ encourages the head clusters (i.e., the clusters with a large proportion of instances) to have larger $b_j$ and the tail clusters (i.e., the clusters with a small proportion of instances) to have smaller $b_j$.
When $b_j$ of a tail cluster approaches 0, 
this tail cluster tends to disappear (clustering degeneracy).
Whereas $\Psi(\bm{b})$ still encourages $\bm{b}$ to be a uniform distribution for avoiding the degeneracy.
With a decent trade-off parameter $\epsilon_2$, SAOT can explore appropriate $\bm{b}$ and obtain reliable $\bm{\pi}$ for various imbalanced data.
We provide the optimization details in Appendix \ref{sec:appsaot}.
After obtaining $\bm{\pi}$, we can get pseudo-labels by argmax operation, i.e,
\begin{equation}
    \begin{aligned}
    \bm{Q}_{ij}=\begin{cases}
        1,&\text{if } j=\underset{j'}{\mathrm{argmax}} \bm{\pi}_{ij'}\\
        0,&\text{otherwise}.
        \end{cases}
    \end{aligned}
\end{equation}
It should be noted that, for convenience, we let $\bm{\pi}=\frac{1}{N}\bm{Q}$ before. However, $\bm{\pi}$ is essentially a join probability matrix and $\pi_{ij}$ can be decimals, while each row of $\bm{Q}$ is a one-hot vector.

Through the steps of cluster assignment and self-adaptive optimal transport, we can generate reliable pseudo-labels on various imbalanced data for the robust representation learning module.

\subsection{Robust Representation Learning module}
\label{class}
We then introduce the robust representation learning module.
To begin with, motivated by \cite{wenzel2022assaying}, we propose to adopt instance augmentations to improve the model robustness against various noises.
Furthermore, inspired by \cite{chen2020simple}, \cite{zhang2021supporting} and \cite{dong2022cml}, we adopt both class-wise and instance-wise contrastive learning to utilize the pseudo-labels and the augmented instance pairs for robust representation learning, as shown in Fig.\ref{fig:model}(b).
The class-wise contrastive learning uses pseudo-labels as the supervision to pull the representations from the same cluster together and push away different clusters.
While the instance-wise contrastive learning disperses different instances apart, which is supposed to separate the overlapped clusters.

Next, we provide the details of the robust representation learning module.
We utilize contextual augmenter \cite{kobayashi2018contextual, ma2019nlpaug} to generate augmented pairs of the original texts as $\bm{X}^{(1)}$ and $\bm{X}^{(2)}$.
Like the cluster assignment step in the pseudo-labels generation module, we can obtain the representations of augmented pairs $\bm{X}^{(1)}$ and $\bm{X}^{(2)}$ as $\bm{E}^{(1)}\in\mathbbm{R}^{N\times D_1}$ and $\bm{E}^{(2)}\in\mathbbm{R}^{N\times D_1}$, respectively.
We can obtain the predictions of them as $\bm{P}^{(1)}\in\mathbbm{R}^{N\times C}$ and $\bm{P}^{(2)}\in\mathbbm{R}^{N\times C}$, respectively.
We use the fully connected layers as the projecting network $G_z$ to map the representations to the space where instance-wise contrastive loss is applied, i.e., $G_z(\bm{E}^{(1)})=\bm{Z}^{(1)}\in\mathbbm{R}^{N\times D_2}$ and $G_z(\bm{E}^{(2)})=\bm{Z}^{(2)}\in\mathbbm{R}^{N\times D_2}$, where $D_2$ is the dimension of the projected representations.
The encoding network and the clustering network share weights with the pseudo-label generation module.

The class-wise contrastive learning enforces consistency between cluster predictions of positive pairs.
Specifically, the two augmentations from the same original text are regarded as a positive pair and the contrastive task is defined on pairs of augmented texts.
Moreover, the pseudo-label of an original text is considered as the target of corresponding two augmented texts.
We use the augmented texts with the targets as supervised data for cross-entropy minimization to achieve the consistency.
The class-wise contrastive loss is defined as below:
\begin{equation}
\label{eq:classification_loss}
    \begin{aligned}
        \mathcal{L}_{C}={\frac{1}{N}}\langle \bm{Q},-\log\bm{P}^{(1)} \rangle +  {\frac{1}{N}}\langle \bm{Q},-\log\bm{P}^{(2)} \rangle.
    \end{aligned}
\end{equation}
The instance-wise contrastive learning enforces consistency between projected representations of positive pairs while maximizing the distance between negative pairs.
Specifically, for a batch, there are $2N$ augmented texts, their projected representations are $\bm{Z}=[\bm{Z}^{(1)},\bm{Z}^{(2)}]^T$, given a positive pair with two texts which are augmented from the same original text, the other $2(N-1)$ augmented texts are treated as negative samples.
The loss for a positive pair $(i, j)$ is defined as:
\begin{equation}
\label{eq:insttance-cl}
    \begin{aligned}
        \mathcal{l}{(i,j)} = 
        -\log{\frac{\exp(\text{sim}(\bm{Z}_i, \bm{Z}_j)/\tau)}{\sum_{k=1}^{2N}\mathbbm{1}_{k\neq i}{\exp(\text{sim}(\bm{Z}_i, \bm{Z}_k)/\tau)}}},
    \end{aligned}
\end{equation}
where $\text{sim}(\bm{u}, \bm{v})$ denotes cosine similarity between $\bm{u}$ and $\bm{v}$, $\tau$ denotes the temperature parameter, and $\mathbbm{1}$ is an indicator. 
The instance-wise contrastive loss is computed across all positive pairs in a batch, including both $(i, j)$ and $(j, i)$. 
That is,
\begin{equation}
    \begin{aligned}
        \mathcal{L}_{I} = \frac{1}{2N}\sum_{i=1}^N({\mathcal{l}{(i, 2i)}} + \mathcal{l}{(2i, i)}).
    \end{aligned}
\end{equation}
By combining the pseudo-supervised class-wise contrastive learning and the instance-wise contrastive learning, we can obtain robust representations and correctly separated clusters.

\subsubsection{Putting Together}
The total loss of \modelname~could be obtained by combining the pseudo-supervised class-wise contrastive loss and the instance-wise contrastive loss. That is, the loss of \modelname~is given as:
\begin{equation}
    \begin{aligned}
        \mathcal{L}=\mathcal{L}_{C}+\lambda_I\mathcal{L}_{I},
    \end{aligned}
\end{equation}
where $\lambda_I$ is a hyper-parameter to balance the two losses. By doing this, \modelname~not only provides robustness to the imbalance in data, but also improve robustness against the noise in data.

The whole model with two modules forms a closed loop and self evolution, which indicates that the learned representations (more robust) and cluster predictions (more accurate) elevate each other progressively, as more reliable pseudo-labels are discovered during the iterations.
Specifically, we firstly initialize the pseudo-labels $\bm{Q}$ by performing k-means on text representations.
Next, we train the robust representation learning module by batch with the supervision of pseudo-labels. 
Meanwhile, we update $\bm{Q}$ throughout the whole training process in a logarithmic distribution, following \cite{asano2020self}.
Finally, we can obtain the cluster assignments by the column index of the largest entry in each row of $\bm{P}$.
The training stops if the change of cluster assignments between two consecutive updates for $\bm{P}$ is less than a threshold $\delta$ or the maximum number of iterations is reached.
%

\section{Experiment}
\begin{table*}
\centering
  \footnotesize
  \begin{tabular}{l
  p{1.00cm}<{\centering}
  p{0.90cm}<{\centering}
  p{0.90cm}<{\centering}
  p{0.90cm}<{\centering}
  p{0.90cm}<{\centering}
  p{0.90cm}<{\centering}
  p{0.90cm}<{\centering}
  p{0.90cm}<{\centering}}
    \toprule
     &\multicolumn{2}{c}{\textbf{AgNews}} &\multicolumn{2}{c}{\textbf{SearchSnippets}} &\multicolumn{2}{c}{\textbf{Stackoverflow}} &\multicolumn{2}{c}{\textbf{Biomedical}}\\
     \cmidrule{2-9}
     & \text{ACC} & \text{NMI} & \text{ACC} & \text{NMI} & \text{ACC} & \text{NMI}& \text{ACC} & \text{NMI}\\
    \midrule
    \text{BOW} & 28.71 & 4.07 & 23.67 & 9.00 & 17.92 & 13.21 & 14.18 & 8.51\\
    \text{TF-IDF} & 34.39 & 12.19 & 30.85 & 18.67 & 58.52 & 59.02 & 29.13 & 25.12 \\
    \text{{STC}$^2$-LPI} & - & - & 76.98 & 62.56 & 51.14 & 49.10 & \underline{43.37} & 38.02 \\
    \text{Self-Train} & - & - & 72.69 & 56.74 & 59.38 & 52.81 & 40.06 & 34.46 \\
    \text{K-means\_IC} & 66.30 & 42.03 & 63.84 & 42.77 & \underline{74.96} & \underline{70.27} & 40.44 & 32.16 \\
    \text{SCCL} & \underline{83.10} & \underline{61.96} & \underline{79.90} & \underline{63.78} & 70.83 & {69.21} & 42.49 & \underline{39.16} \\
    \midrule
    \text{SBERT(k-means)}  & 65.95 & 31.55 & 55.83 & 32.07 & 60.55 & 51.79 & 39.50 & 32.63\\
    \text{\modelname-OT} & 65.94 & 41.86 & 70.79 & 59.30 & 56.77 & 60.17 & 38.14 & 34.89\\
    \text{\modelname-C} & 78.08 & 49.39 & 62.59 & 44.02 & 78.33 & 70.28 & 46.74 & 38.69 \\
    \text{\modelname-I} & \textbf{85.39} & \textbf{62.79} & 79.26 & 68.03 & 31.31 & 28.66 & 34.39 & 31.20 \\
    \text{\modelname}  & {84.24} & {62.45} & \textbf{80.10} & \textbf{69.74} & \textbf{83.30} & \textbf{74.11} & \textbf{48.40} & \textbf{40.12}\\
    \textbf{Improvement($\uparrow$)} & \textbf{1.14} & \textbf{0.49} & \textbf{0.20} & \textbf{5.96} & \textbf{8.34} & \textbf{3.84} & \textbf{5.03} & \textbf{0.96} \\
    \bottomrule
  \end{tabular}
  \begin{tabular}{l
  p{1.00cm}<{\centering}
  p{0.90cm}<{\centering}
  p{0.90cm}<{\centering}
  p{0.90cm}<{\centering}
  p{0.90cm}<{\centering}
  p{0.90cm}<{\centering}
  p{0.90cm}<{\centering}
  p{0.90cm}<{\centering}}
     &\multicolumn{2}{c}{\textbf{GoogleNews-TS}} &\multicolumn{2}{c}{\textbf{GoogleNews-T}} &\multicolumn{2}{c}{\textbf{GoogleNews-S}} &\multicolumn{2}{c}{\textbf{Tweet}}\\
     \cmidrule{2-9}
     & \text{ACC} & \text{NMI} & \text{ACC} & \text{NMI} & \text{ACC} & \text{NMI}& \text{ACC} & \text{NMI}\\
    \midrule
    \text{BOW} & 58.79 & 82.59 & 48.05 & 72.38 & 52.68 & 76.11 & 50.25 & 72.00 \\
    \text{TF-IDF} & 69.00 & 87.78 & 58.36 & 79.14 & 62.30 & 83.00 & 54.34 & 78.47 \\
    \text{K-means\_IC}  & 79.81 & 92.91 & 68.88 & 83.55 & \underline{74.48} & \underline{88.53} & 66.54 & 84.84\\
    \text{SCCL} & \underline{82.51} & \underline{93.01} & \underline{69.01} & \underline{85.10} & 73.44 & 87.98 & \underline{73.10} & \underline{86.66} \\
    \midrule
    \text{SBERT(k-means)}  & 65.71 & 86.60 & 55.53 & 78.38 & 56.62 & 80.50 & 53.44 & 78.99\\
    \text{\modelname-OT}  & 63.97 & 85.79 & 56.45 & 79.49 & 59.48 & 81.21 & 56.84 & 79.16\\
    \text{\modelname-C} & 78.48 & 90.59 & 63.08 & 82.16 & 65.05 & 83.88 & \textbf{76.62} & 85.61 \\
    \text{\modelname-I} & 75.44 & 92.06 & 64.84 & 85.06 & 66.22 & 86.93 & 61.12 & 84.53\\
    \text{\modelname} & \textbf{83.27} & \textbf{93.15} & \textbf{72.27} & \textbf{87.39} & \textbf{79.32} & \textbf{89.40} & 75.20 & \textbf{87.35} \\
    \textbf{Improvement($\uparrow$)} & \textbf{0.76} & \textbf{0.14} & \textbf{3.26} & \textbf{2.29} & \textbf{4.84} & \textbf{0.87} & \textbf{2.1} & \textbf{0.69} \\
    \bottomrule
  \end{tabular}
  \vspace{-0.2cm}
\caption{Experimental results on eight short text datasets. We bold the \textbf{best result}, underline the \underline{runner-up}.}
\vspace{-0.4cm}
\label{ta:result}
\end{table*}

In this section, we conduct experiments on several real-world datasets to answer the following questions: 
(1) \textbf{RQ1}: How does our approach perform compared with the state-of-the-art short text clustering methods? 
(2) \textbf{RQ2}: How do the SAOT, and the two contrastive losses contribute to the performance improvement? 
(3) \textbf{RQ3}: How does the performance of \modelname~vary with different values of the hyper-parameters?

\subsection{Datasets}
We conduct extensive experiments on eight popularly used real-world datasets, i.e., \textbf{AgNews}, \textbf{StackOverflow}, \textbf{Biomedical}, \textbf{SearchSnippets}, \textbf{GoogleNews-TS},
\textbf{GoogleNews-T}, \textbf{GoogleNews-S} and \textbf{Tweet}.
Among them, \textbf{AgNews}, \textbf{StackOverflow} and \textbf{Biomedical} are balanced datasets, \textbf{SearchSnippets} is a light imbalanced dataset, \textbf{GoogleNews}, \textbf{GoogleNews-T}, \textbf{GoogleNews-S} and \textbf{Tweet} are heavy imbalanced datasets.
Following \cite{zhang2021supporting}, we take unpreprocessed data as input to demonstrate that our model is robust to noise, for a fair comparison.
More details about the datasets are shown in Appendix \ref{sec:appdata}.

\subsection{Experiment Settings}
\label{sec:implementation}
We build our model with PyTorch \cite{NEURIPS2019_9015} and train it using the Adam optimizer \cite{kingma2014adam}.
We study the effect of hyper-parameters $\epsilon_1$ and $\epsilon_2$ on SAOT by varying them in $\{0.05,0.1,0.2,0.5\}$ and $\{0,0.001,0.01,0.1,1\}$, respectively.
Besides, we study the effect of the hyper-parameter $\lambda_I$ by varying it in $\{0,1,5,10,20,50,100\}$.
The more details are provided in Appendix \ref{sec:appimple}.
Following previous work \cite{xu2017self,hadifar2019self,rakib2020enhancement,zhang2021supporting}, we set the cluster numbers to the ground-truth category numbers, and we adopt Accuracy (ACC) and Normalized Mutual Information (NMI) to evaluate different approaches.
The specific definitions of the evaluation methods are shown in Appendix \ref{sec:appeval}.
For all the experiments, we repeat five times and report the average results.

\subsection{Baselines}
We compare our proposed approach with the following short text clustering methods.
\textbf{BOW} \cite{scott1998text} \& \textbf{TF-IDF} \cite{salton1983introduction} applies k-means on the TF-IDF representations and BOW representations respectively.
\textbf{STC$^2$-LPI} \cite{xu2017self} first uses word2vec to train word embeddings on the in-domain corpus, and then uses a convolutional neural network to obtain the text representations where k-means is applied for clustering.
\textbf{Self-Train} \cite{hadifar2019self} follows \cite{xie2016unsupervised} uses an autoencoder to get the representations, and finetunes the encoding network with the same clustering objective.
The difference are that it uses the word embeddings provided by \cite{xu2017self} with SIF \cite{arora2017simple} to enhance the pretrained word embeddings, and obtains the final cluster assignments via k-means.
\textbf{K-means\_IC} \cite{rakib2020enhancement} first applies k-means on the TF-IDF representations and then enhances clustering by the iterative classification algorithm.
\textbf{SCCL} \cite{zhang2021supporting} is the more recent short text clustering model which utilizes SBERT \cite{reimers2019sentence} as the backbone and introduces instance-wise contrastive learning to support clustering.
Besides, \textbf{SCCL} uses the clustering objective proposed in \cite{xie2016unsupervised} for deep joint clustering and  obtains the final cluster assignments by k-means.

\subsection{Clustering Performance (RQ1)}
\paragraph{Results and discussion} The comparison results on eight datasets are shown in Table \ref{ta:result}. 
\textbf{SBERT(k-means)} denotes the pre-trained SBERT model with k-means clustering, which is the initial state of our \modelname.

From the results, we can find that: 
(1) Only adopting traditional text representations (\textbf{BOW} and \textbf{TF-IDF}) cannot obtain satisfying results due to the data sparsity problem.
(2) Deep learning methods (\textbf{STC$^2$-LPI} and \textbf{Self-Train}) outperform traditional ones, indicating that the application of pre-trained word embeddings and deep neural network alleviates the sparsity problem.
(3) \textbf{SCCL} obtains better results by introducing instance-wise contrastive learning to cope with the noises problem.
However, the clustering objective of \textbf{SCCL} is prone to obtain degenerate solutions \cite{yang2017towards,ji2019invariant}.
Besides, it is suboptimal for extra application of k-means \cite{van2020scan}.
The degeneracy gives the representation learning wrong guidance, which degrades the final k-means clustering performance.
(4) \modelname~outperforms all baselines, which proves that the robust representation learning with pseudo-supervised class-wise contrastive learning and instance-wise contrastive learning can significantly improve the clustering performance.
To better show the clustering degeneracy problem, we visualize how the number of predicted clusters are changing over iterations on \textbf{SCCL} and \modelname~in Appendix \ref{sec:appvisual}.
From the results, we verify that \textbf{SCCL} has relatively serious clustering degeneracy problem while \modelname~solves it.
The visualization results illustrate the validity of our model.
\begin{figure*}[t]
  \centering
    \subfigure[SBERT]{
    \begin{minipage}[t]{0.23\linewidth} 
    \includegraphics[width=4.1cm]{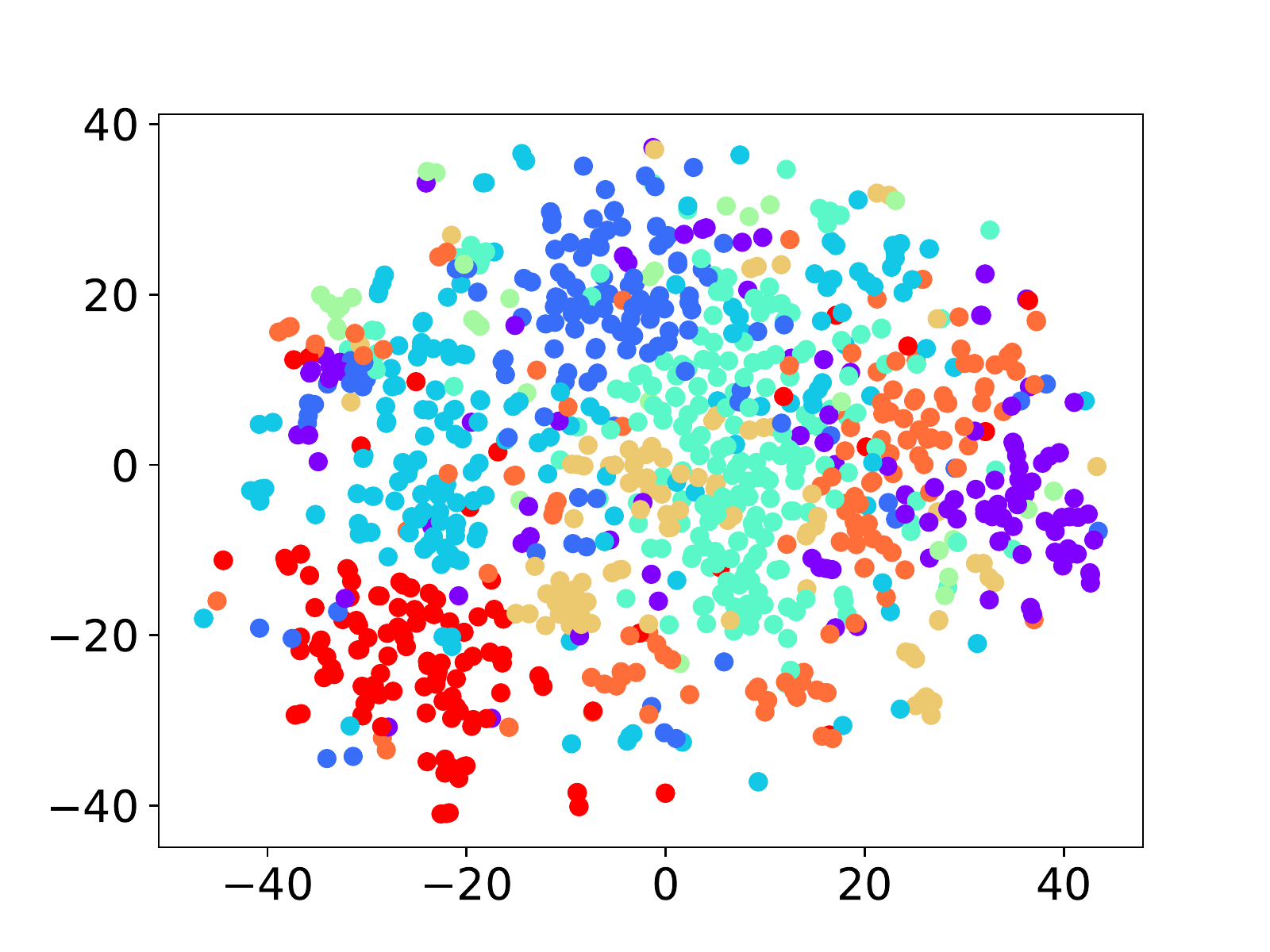}
    \end{minipage}
    }    
    \subfigure[\modelname-C]{
    \begin{minipage}[t]{0.23\linewidth} 
    \includegraphics[width=4.1cm]{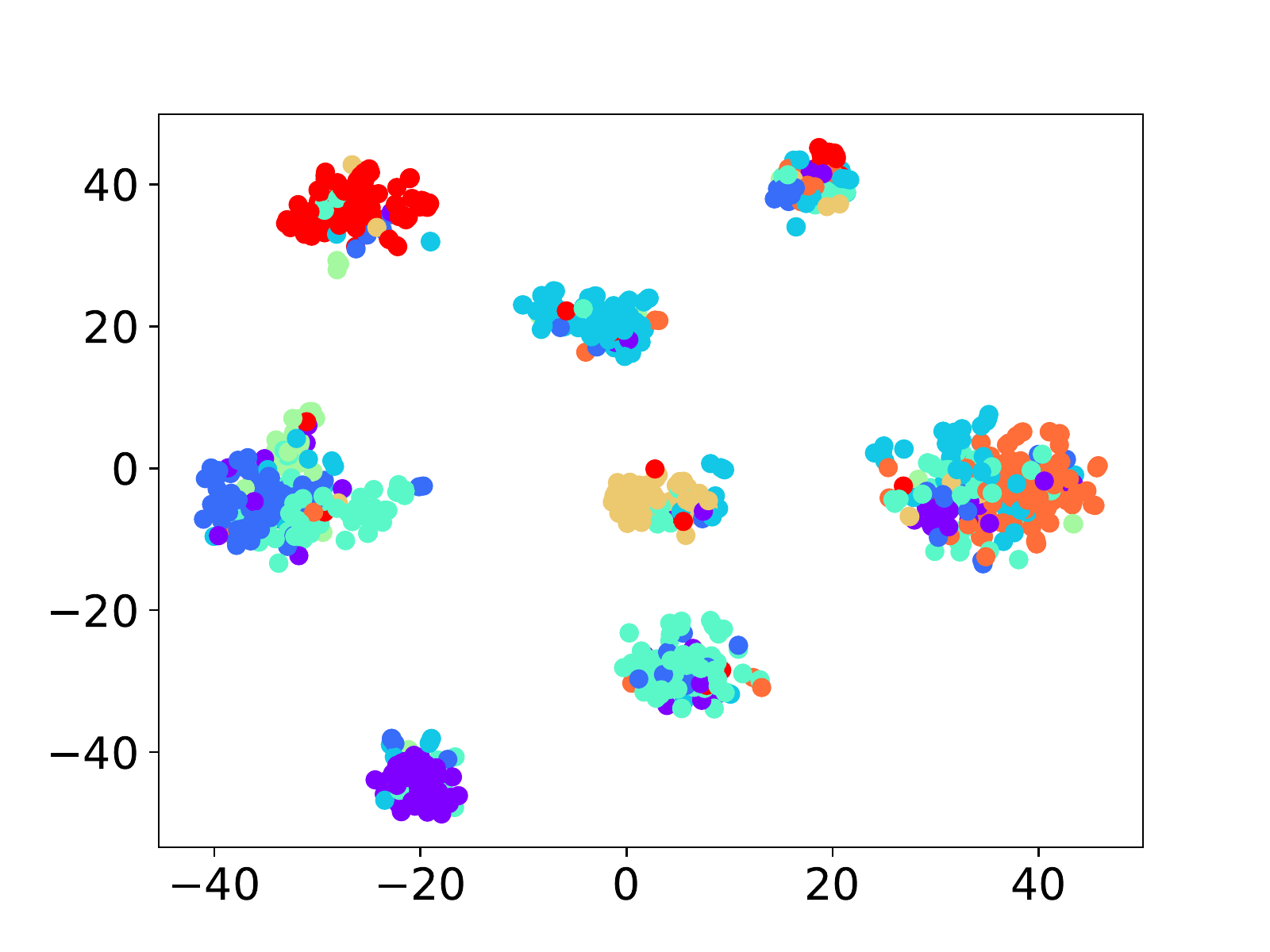}
    \end{minipage}
    }
    \subfigure[\modelname-I]{
    \begin{minipage}[t]{0.23\linewidth} 
    \includegraphics[width=4.1cm]{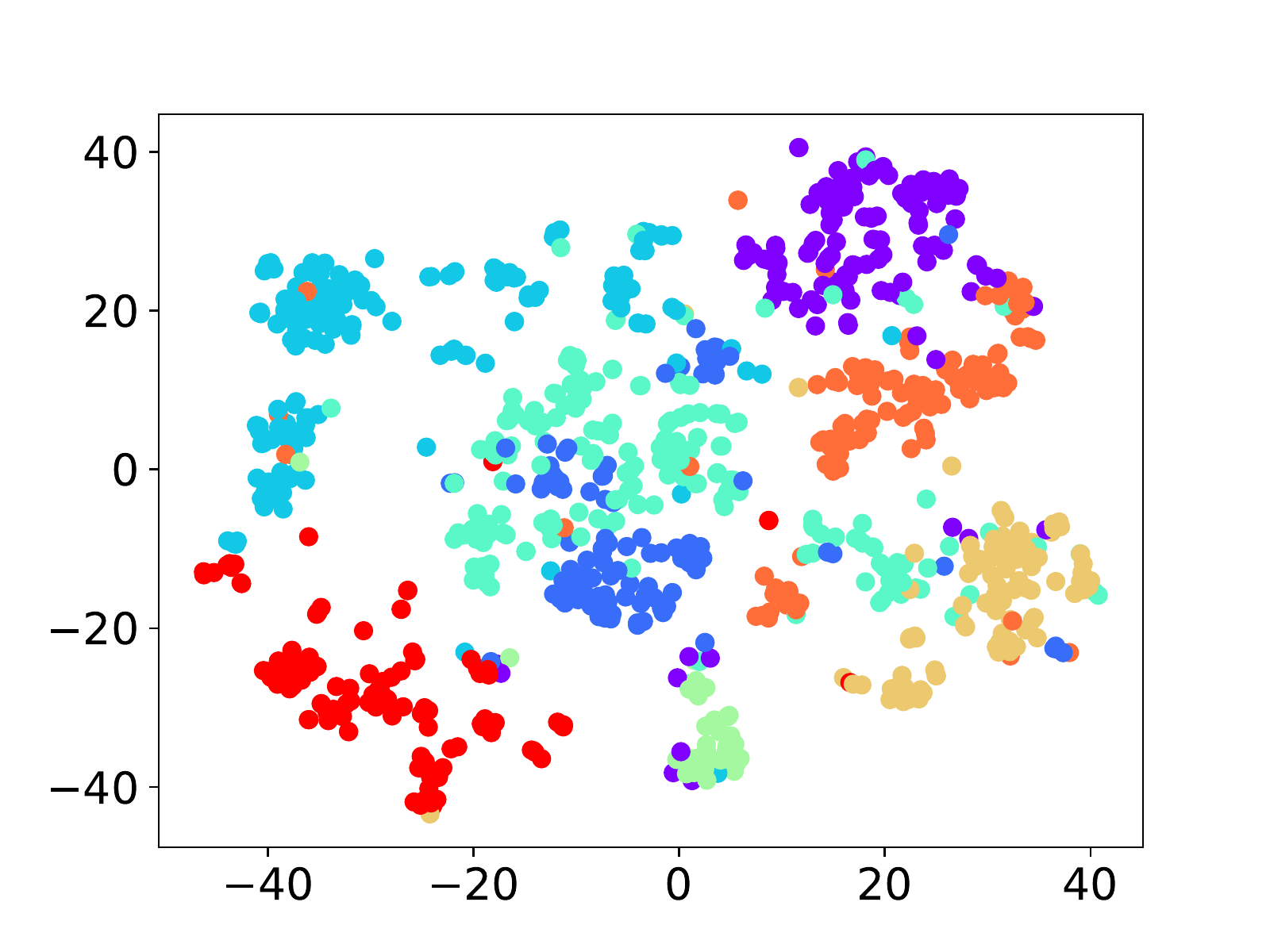}
    \end{minipage}
    }
    \subfigure[\modelname]{
    \begin{minipage}[t]{0.23\linewidth} 
    \includegraphics[width=4.1cm]{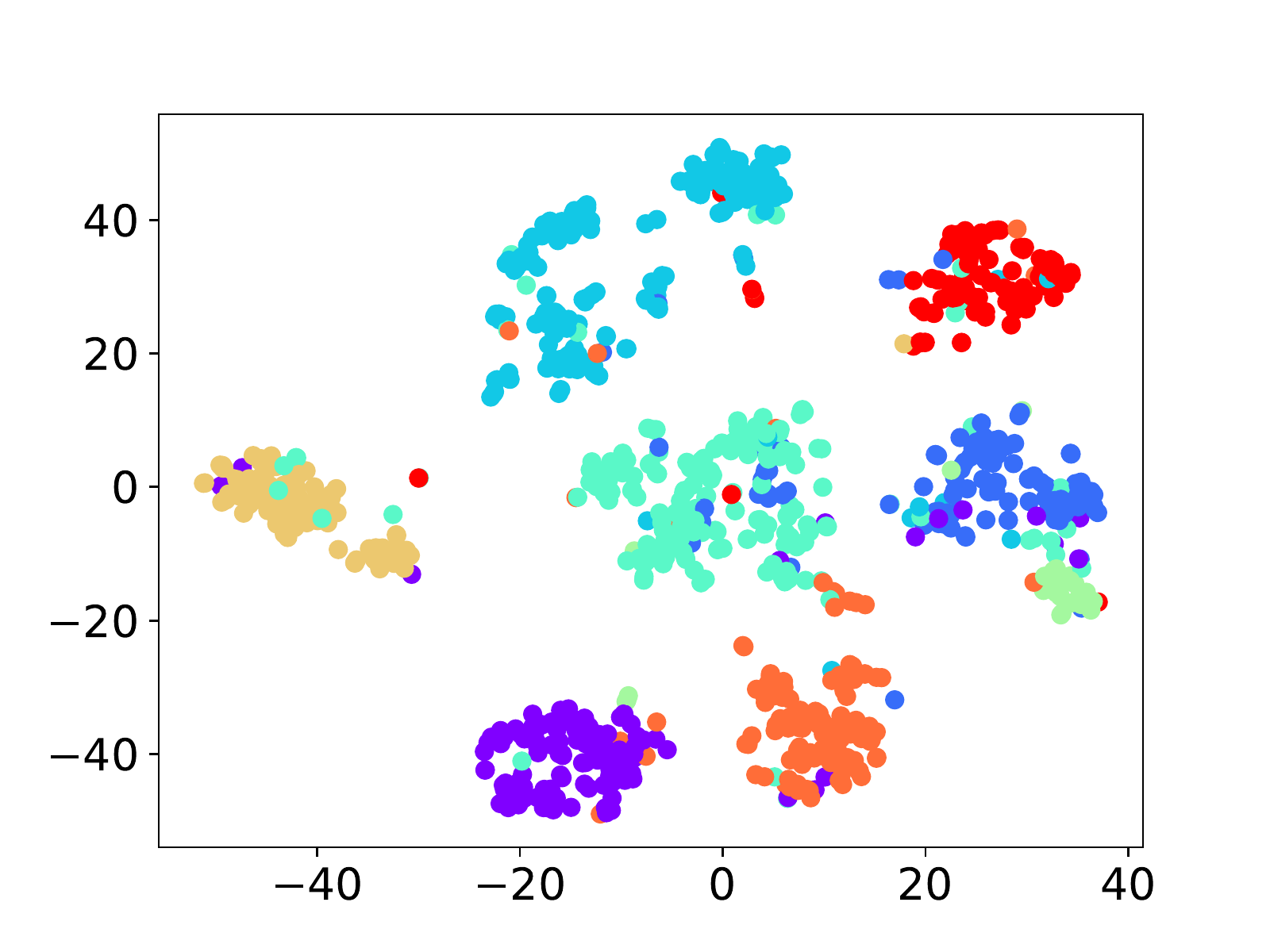}
    \end{minipage}
    }
    \vspace{-0.4cm}
  \caption{TSNE visualization of the representations on SearchSnippets, each color indicates a ground truth category.}
  \vspace{-0.4cm}
  \label{fig:tsne}
\end{figure*}

\subsection{In-depth Analysis (RQ2 and RQ3)}
\paragraph{Ablation (RQ2)} To study how each component of \modelname~contribute to the final performance, we compare \modelname~with its several variants, including \modelname-OT, \modelname-C and \modelname-I. 
\modelname-OT adopts both the pseudo-supervised class-wise contrastive learning and instance-wise contrastive learning, while the pseudo-labels are generated by traditional OT with fixed random class distribution.
\modelname-C only adopts the pseudo-supervised class-wise contrastive learning, the pseudo-labels are generated by SAOT.
\modelname-I only adopts the instance-wise contrastive learning while the clustering results are obtained by k-means.

The comparison results are shown in Table \ref{ta:result}.
From it, we can observe that they all cannot achieve satisfactory results due to their limitations.
Specifically,
(1) \modelname-OT will be guided by the mismatched distribution constraint to generate unreliable pseudo-labels.
(2) \modelname-C is good at aggregating instances, but it has difficulties to address the situation when different categories are overlapped with each other in the representation space at the beginning of the learning progress, which may lead to a false division.
(3) \modelname-I is good at dispersing different instances, but it has limited ability to aggregate instances which may lead to the unclear boundaries between clusters.
(4) \modelname~achieves the best performance with the combination of pseudo-supervised class-wise contrastive learning and instance-wise contrastive learning while the pseudo-labels are generated by SAOT.
Overall, the above ablation study demonstrates that our proposed SAOT and robust representation learning are effective in solving the short text clustering problem.

\paragraph{Visualization} To further show the functions and the limitations of the pseudo-supervised class-wise contrastive learning and the instance-wise contrastive learning, we visualize the representations using t-SNE \cite{van2008visualizing} for \textbf{SBERT} (initial state), \modelname-C, \modelname-I and \modelname.
The results on \textbf{SearchSnippets} are shown in Fig.\ref{fig:tsne}(a)-(d).
From it, we can see that: 
(1) \textbf{SBERT} (initial state) has no boundaries between classes, and the points from different clusters have significant overlap.
(2) Although \modelname-C groups the representations to exact eight clusters, a large proportion of points are clustered by mistake.
(3) \modelname-I disperses the overlapped categories to some extent, but the points are not clustered.
(4) With the combination of \modelname-C and \modelname-I, \modelname~obtains best text representations with small intra-cluster distance, large inter-cluster distance while only a small amount of points are clustered wrongly.
%
The reasons for these phenomenons are the same as previous results analyzed in \textbf{Ablation}.
\begin{figure}[t]
  \centering
    \subfigure[Effects of $\epsilon_1$]{
    \begin{minipage}[t]{0.45\linewidth} 
    \includegraphics[width=4cm]{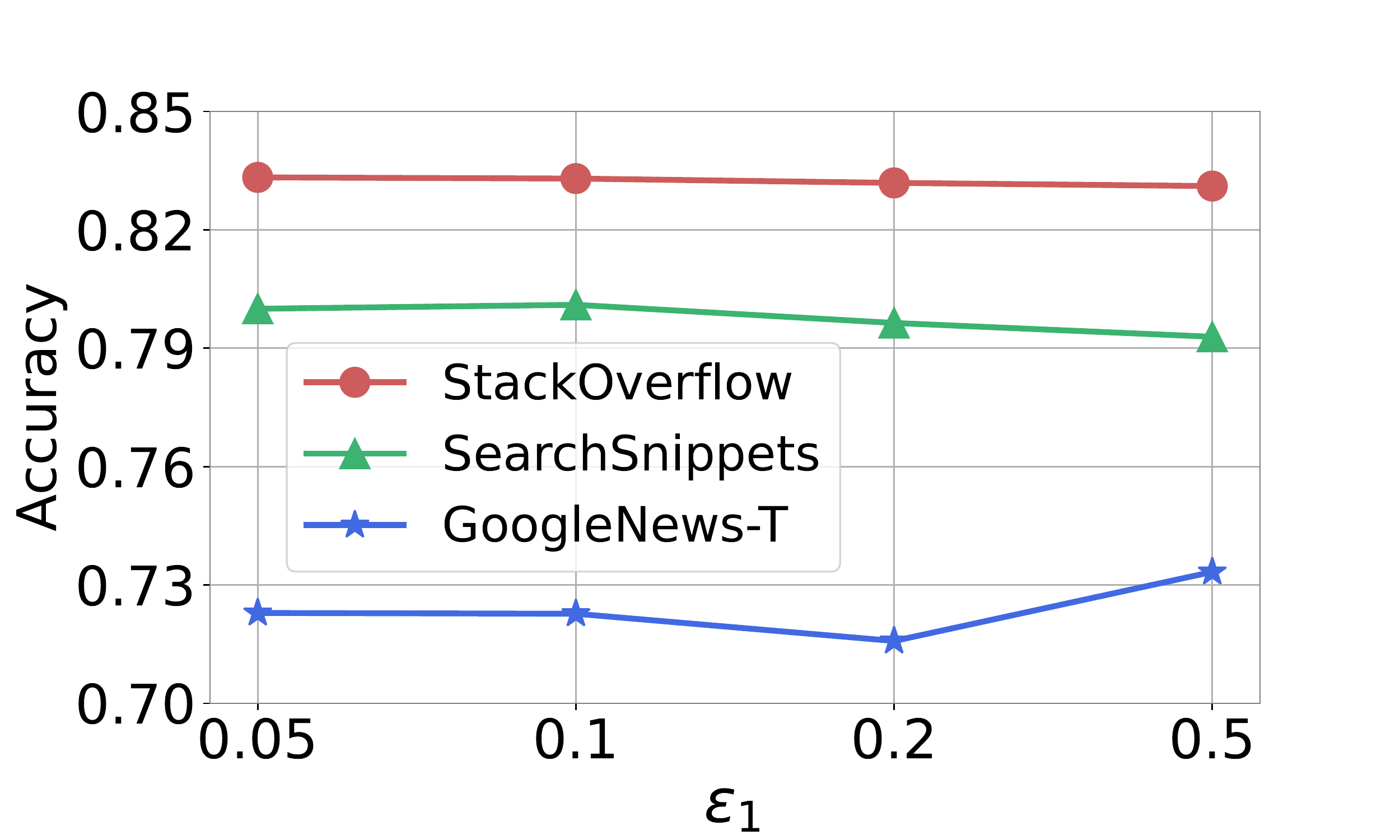}
    \end{minipage}
    }
    \subfigure[Effects of $\epsilon_2$]{
    \begin{minipage}[t]{0.45\linewidth} 
    \includegraphics[width=4cm]{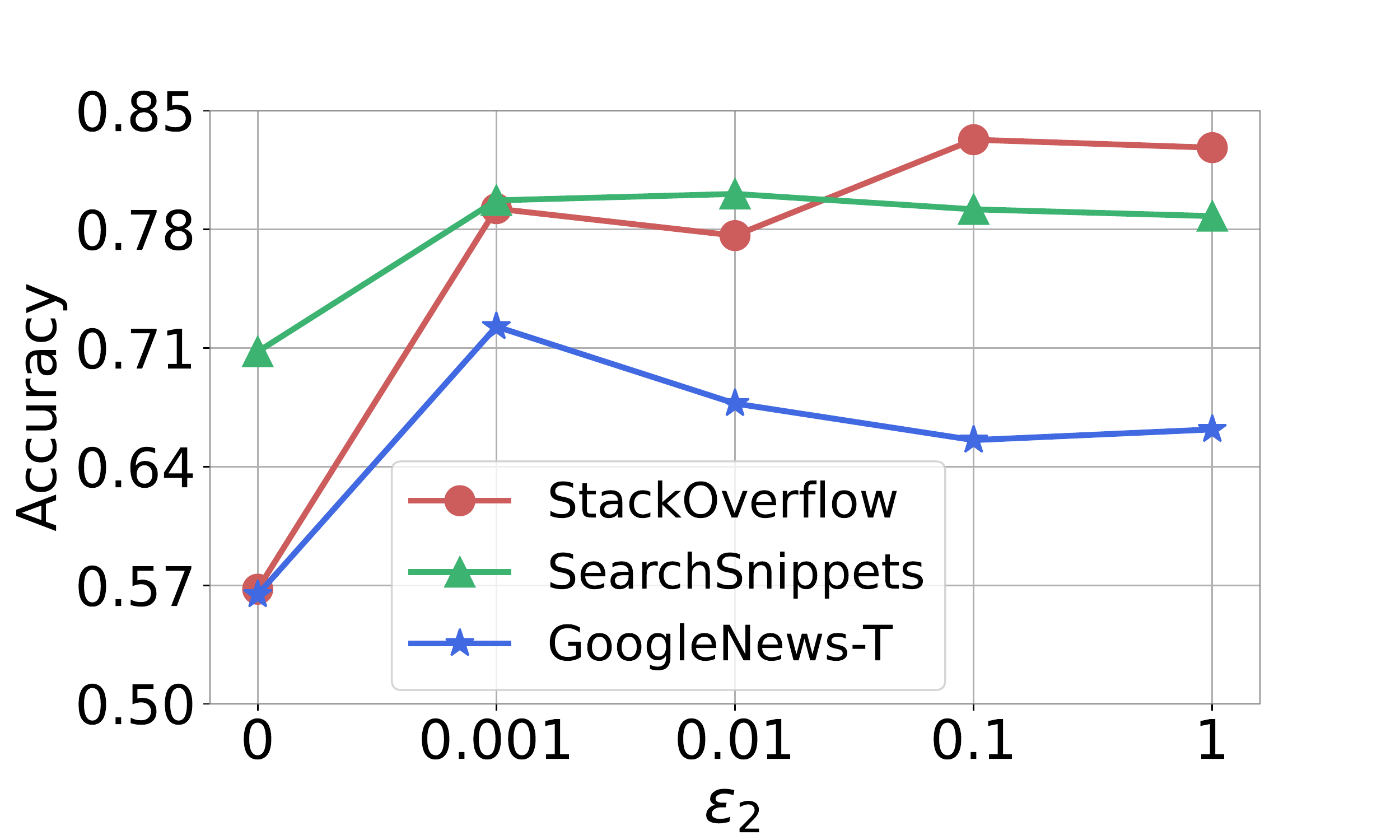}
    \end{minipage}
    }
    \vspace{-0.4cm}
  \caption{The effect of $\epsilon_1$ and $\epsilon_2$ on model accuracy.}
  \vspace{-0.4cm}
  \label{fig:pararotesult}
\end{figure}
\begin{figure}[t]
  \centering
    \subfigure[Effects on ACC]{
    \begin{minipage}[t]{0.45\linewidth} 
    \includegraphics[width=4cm]{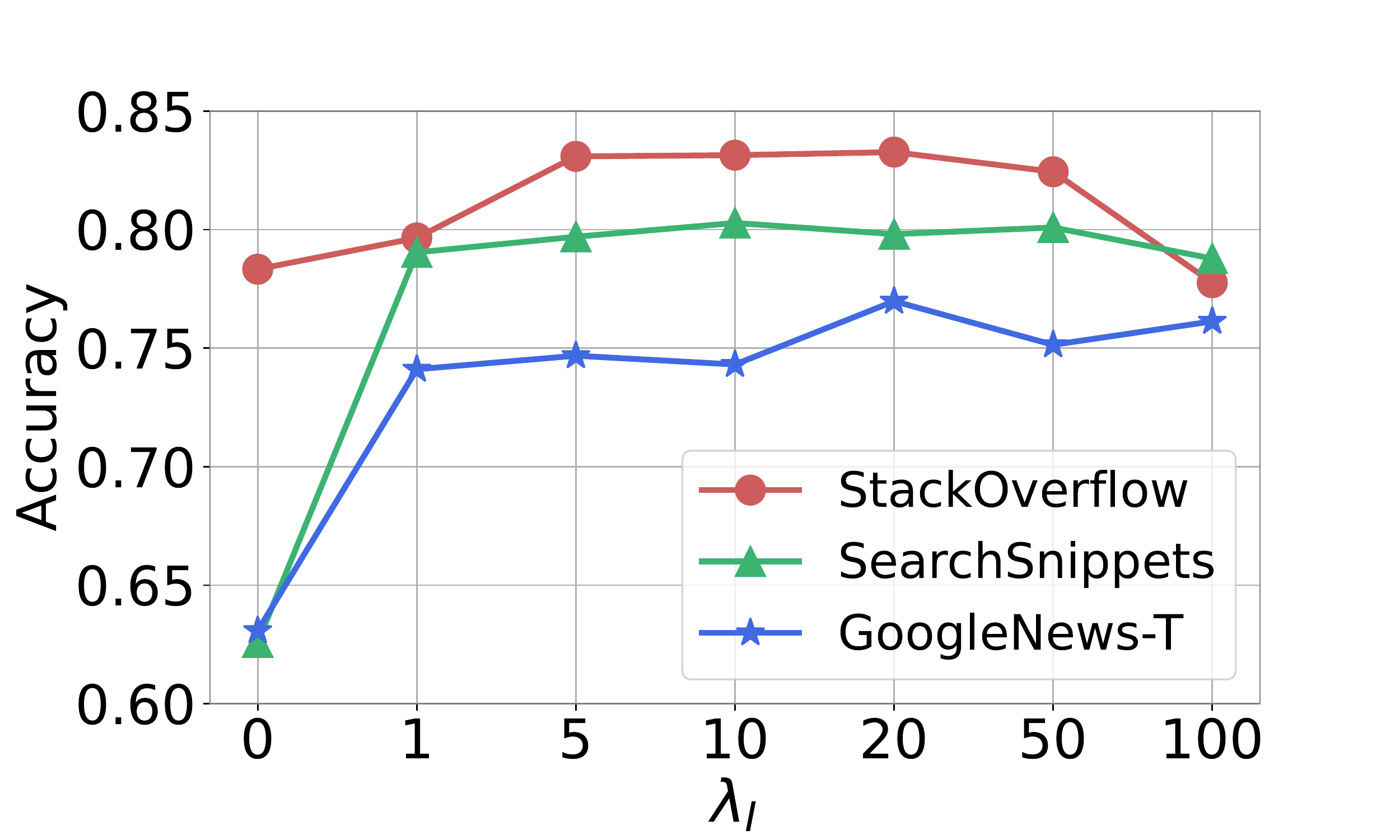}
    \end{minipage}
    }
    \subfigure[Effects on NMI]{
    \begin{minipage}[t]{0.45\linewidth} 
    \includegraphics[width=4cm]{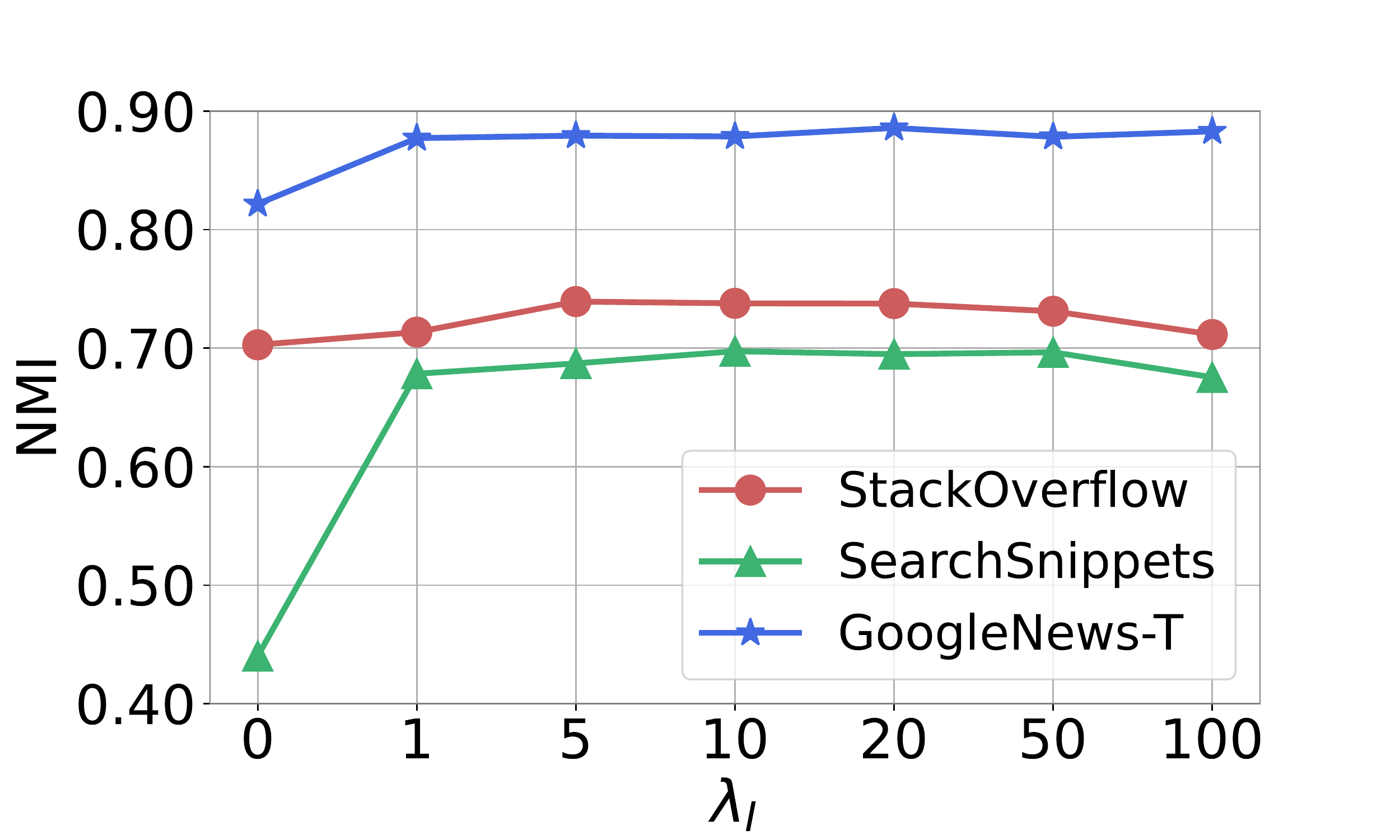}
    \end{minipage}
    }
    \vspace{-0.4cm}
  \caption{The effect of $\lambda_I$ on model performance.}
    \vspace{-0.4cm}
  \label{fig:pararesult}
\end{figure}

\paragraph{Effect of hyper-parameter (RQ3)}
We now study the effects of hyper-parameters on model performance, including $\epsilon_1$, $\epsilon_2$ and $\lambda_I$.
We first study the effects of $\epsilon_1$ and $\epsilon_2$ by varying them in $\{0.05, 0.1, 0.2, 0.5\}$ and $\{0, 0.001, 0.01, 0.1, 1\}$, respectively.
The results are reported in Fig. \ref{fig:pararotesult}(a) and Fig. \ref{fig:pararotesult}(b).
Fig. \ref{fig:pararotesult}(a) shows that the accuracy are
not sensitive to $\epsilon_1$.
Fig. \ref{fig:pararotesult}(b) shows that choosing the proper hyper-parameters for different imbalance levels of datasets is important, especially on the heavy imbalanced dataset \textbf{GoogleNews-T}.
Empirically, we choose $\epsilon_1=0.1$ on all datasets,
$\epsilon_2=0.1$ on the balanced datasets, $\epsilon_2=0.01$ on the light imbalanced datasets, and $\epsilon_2=0.001$ on the heavy imbalanced datasets.
Then we perform experiments by varying $\lambda_I$ in $\{0,1,5,10,20,50, 100\}$.
The results on three datasets are shown in Fig. \ref{fig:pararesult}. 
From them, we can see that the performance improves when $\lambda_I$ increases, then keeps a relatively stable level after $\lambda_I$ reaches $1$ and finally decreases when $\lambda_I$ becomes too large.
We can conclude that when $\lambda_I$ is too small, the ability of instance-wise contrastive learning cannot be fully exploited.
When $\lambda_I$ is too large, the ability of class-wise contrastive learning will be suppressed, which also reduces the clustering performance.
Empirically, we choose $\lambda_I=10$ for all datasets.

\section{Conclusion}
In this paper, we propose a robust short text clustering (\modelname) model, which includes \textit{pseudo-label generation module} and \textit{robust representation learning module}.
The former generates pseudo-labels as the supervision for the latter.
We innovatively propose SAOT in the pseudo-label generation module to provide robustness against the imbalance in data.
We further propose to combine class-wise contrastive learning with instance-wise contrastive learning in the robust representation learning module to provide robustness against the noise in data.
Extensive experiments conducted on eight real-world datasets demonstrate the superior performance of our proposed \modelname.

\section{Limitations}
Like existing short text clustering methods, we assume the real cluster number is known. In the future, we would like to explore a short text clustering method with an unknown number of clusters. Moreover, the time complexity of self-adaptive optimal transport is $O(n^2)$, we are going to seek a new computation to reduce the complexity.

\section*{Acknowledgements}
This work was supported in part by the Leading Expert of “Ten
Thousands Talent Program” of Zhejiang Province (No.2021R52001)
and National Natural Science Foundation of China (No.72192823).

\bibliography{anthology,custom}
\bibliographystyle{acl_natbib}
\appendix
\label{sec:app}
\section{Different Class Distribution Estimation Methods}
\label{sec:apph}
We have tried three class distribution estimation methods, including: (1) \textbf{M1}(ours): The method proposed in our paper with the penalty function $\Psi(\bm{b}=-\log(\bm{b})-\log(1-\bm{b}))$, and $\bm{b}$ can be updated during the process of solving the OT problem.
(2) \textbf{M2}: The method proposed in \cite{wang2022solar} holds no assumption on $\bm{b}$, and $\bm{b}$ can be updated by the model predicted results with moving-average mechanism, that is, $\bm{b}=\mu\hat{\bm{b}}+(1-\mu)\bm{\gamma}$, where $\mu$ is the moving-average parameter, $\bm{\hat{b}}$ is the last updated $\bm{b}$ and ${\gamma}_j=\frac{1}{N}\sum_{i=1}^N\mathbbm{1}(j=\mathop{\arg\max}\bm{P}_i)$. 
(3) \textbf{M3}: This method replaces the penalty function in our method with the common entropy regularization $\Psi(\bm{b})=KL(\bm{b}\parallel\bm{\hat{b}})$, where $\bm{\hat{b}}$ is the last updated $\bm{b}$, and the current $\bm{b}$ can be updated the same way our method does. Note that, the parameters of \textbf{M2} are following \cite{wang2022solar}, the parameters of \textbf{M3} are the same as \textbf{M1}(ours).

\begin{figure}[t]
  \centering
    \subfigure[Accuracy]{
    \begin{minipage}[t]{0.45\linewidth} 
    \includegraphics[width=4cm]{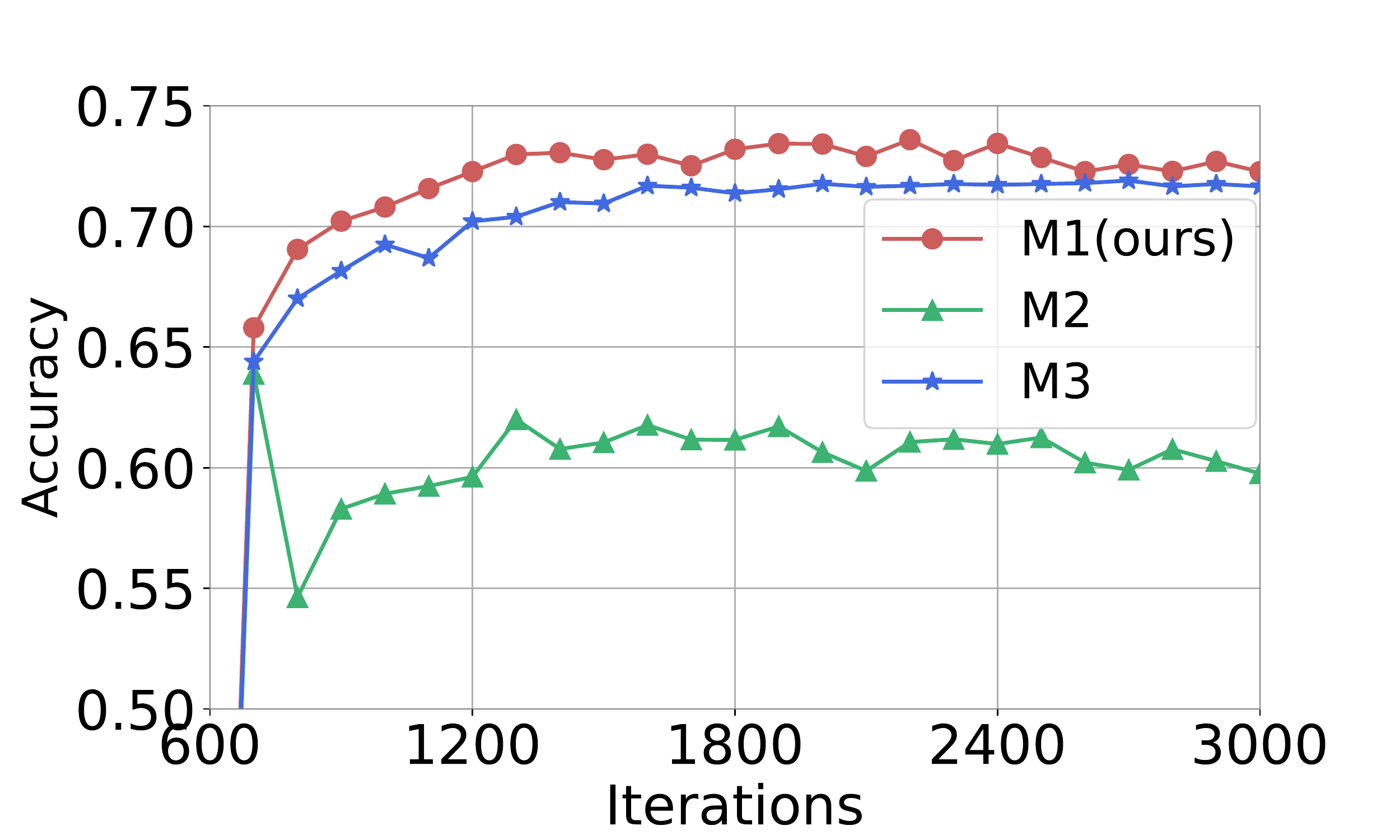}
    \end{minipage}
    }
    \subfigure[Clusters]{
    \begin{minipage}[t]{0.45\linewidth} 
    \includegraphics[width=4cm]{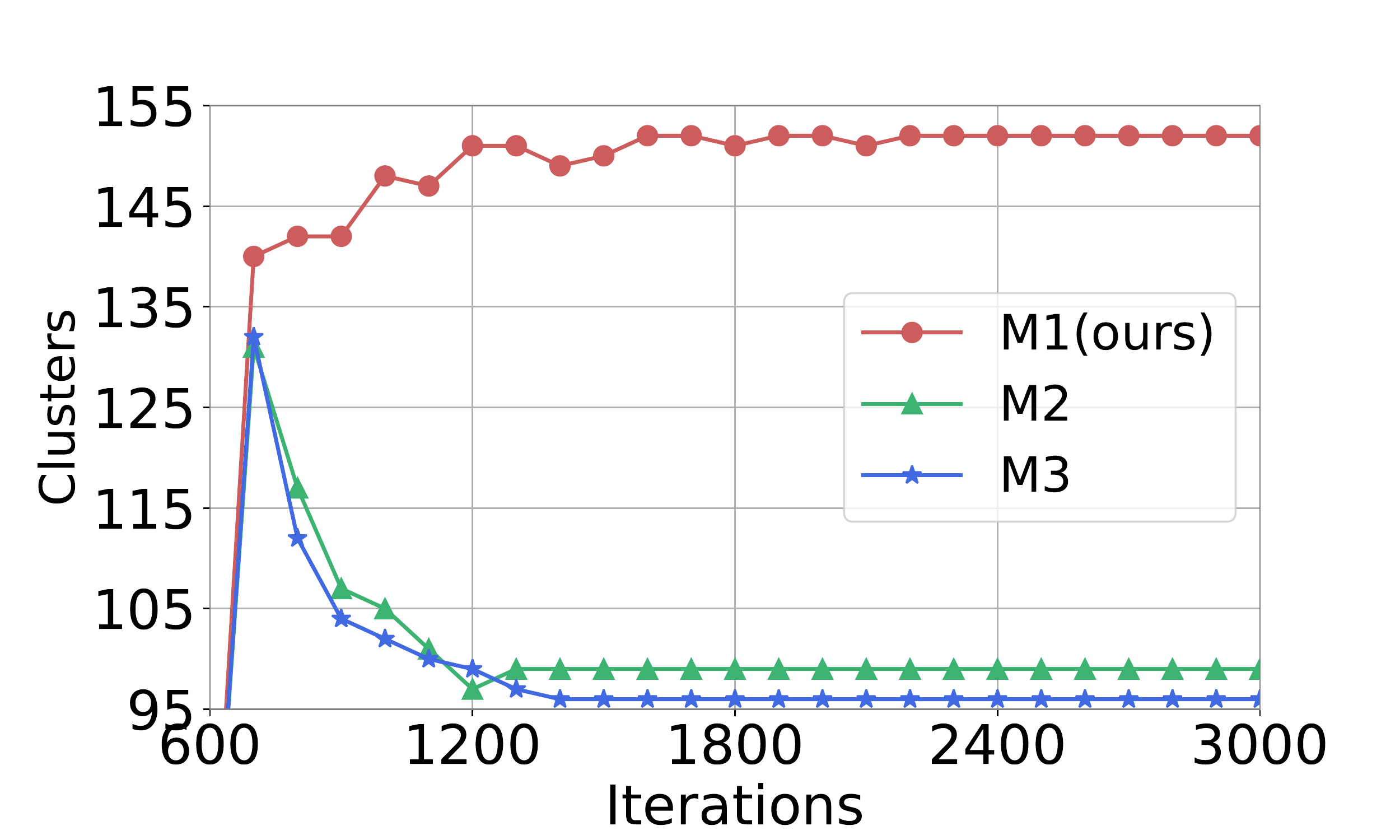}
    \end{minipage}
    }
    \subfigure[Accuracy]{
    \begin{minipage}[t]{0.45\linewidth} 
    \includegraphics[width=4cm]{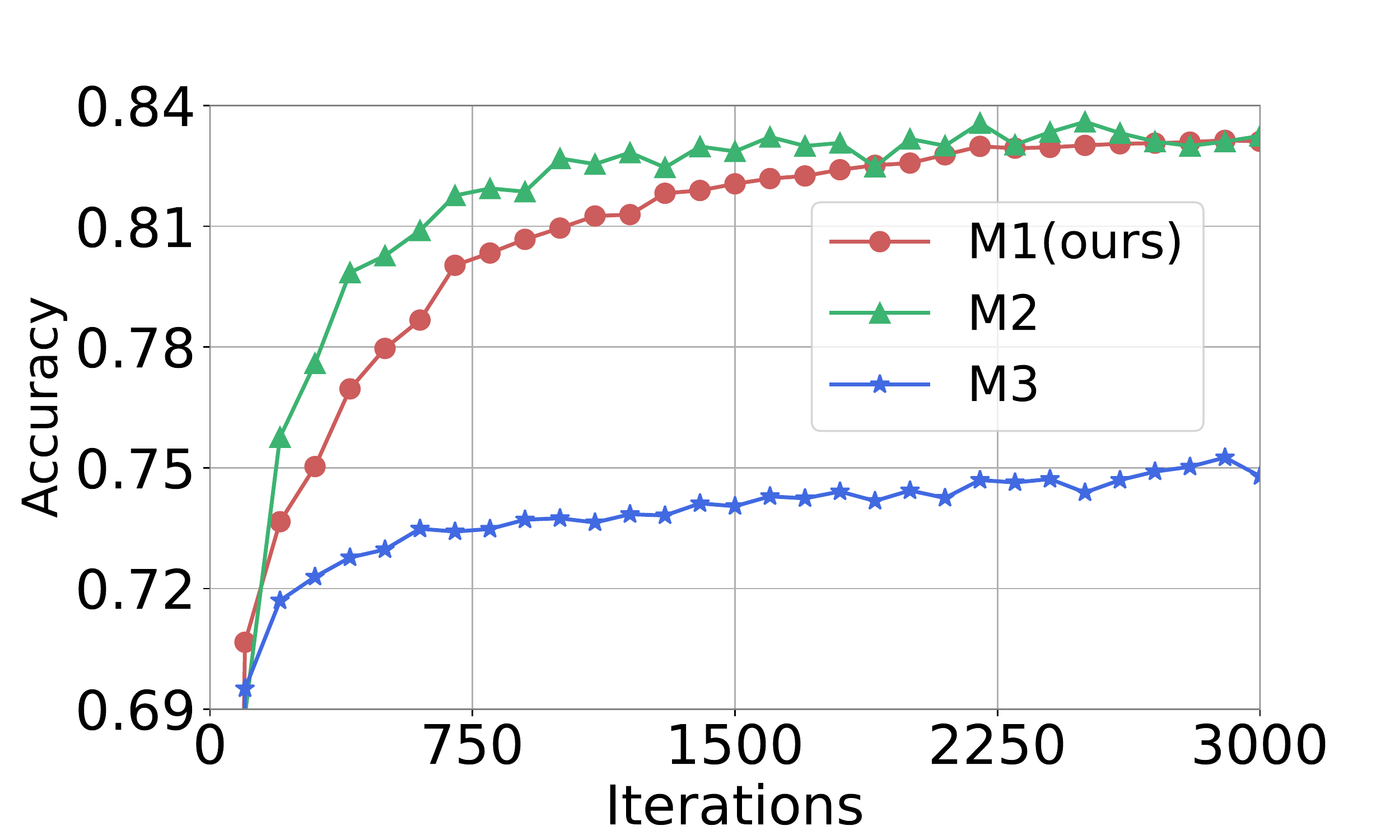}
    \end{minipage}
    }
    \subfigure[Clusters]{
    \begin{minipage}[t]{0.45\linewidth} 
    \includegraphics[width=4cm]{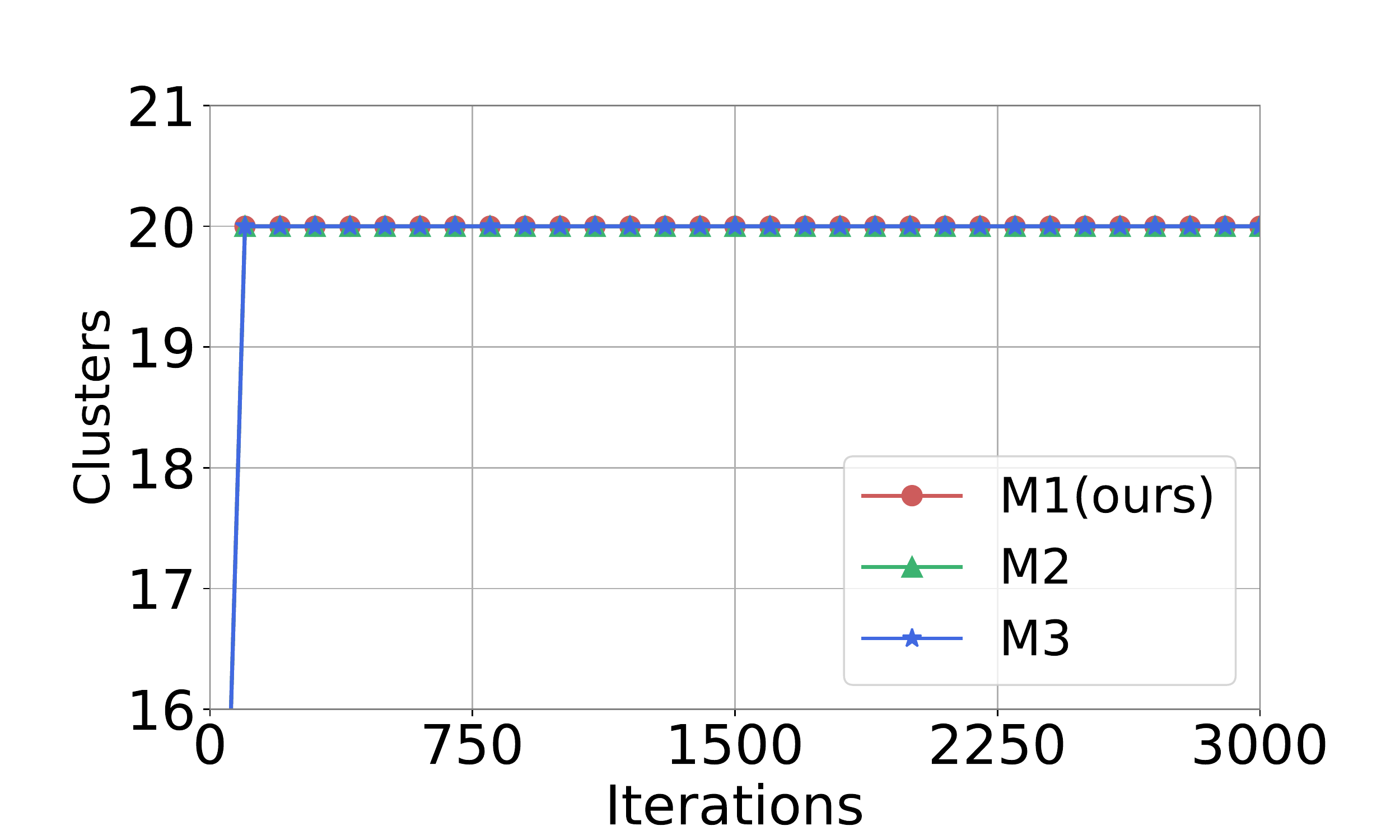}
    \end{minipage}
    }
  \caption{The accuracy and the number of predicted clusters at different iterations on GoogleNews-T (first row) and StackOverflow (second row).
  Note that because the samples in GoogleNews-T are too short, which makes it difficult to generate relatively reliable pseudo-labels,
  we pre-train the representations with $\mathcal{L}_I$ for three methods in the first 600 steps.
  Due to the same pre-training process, we omit the curves in the first 600 steps on GoogleNews-T.
  }
  \label{fig:hresult}
\end{figure}

For comprehensive comparison, we conduct the experiments on one imbalanced dataset \textbf{GoogleNews-T} and one balanced dataset \textbf{StackOverflow} with randomly initialized $\bm{b}$ for visualizing how the accuracy and the number of predicted clusters are changing over iterations. 
Moreover, except the update of $\bm{b}$, everything else about the experiments is the same for three methods.
The results are shown in Fig.\ref{fig:hresult}(a)-(d).
From them, we can find that:
(1) For the imbalanced dataset, \textbf{M1}(ours) achieves the best accuracy and converges to the real category number, while other methods have clustering degeneracy problem.
(2) For the balanced dataset, \textbf{M2} achieves best accuracy more quickly while \textbf{M1}(ours) catches up in the end, and all methods obtain real category number.
Although \textbf{M3} can obtain good accuracy on the imbalanced dataset, it has the worst accuracy on the balanced dataset.
In addition, although \textbf{M2} achieves good accuracy on the balanced dataset, it has the worst accuracy on the imbalanced dataset.
Only \textbf{M1}(ours) achieves fairly good performance on both datasets, which indicates that our method are robust to various imbalance levels of datasets.
The experiments prove the effectiveness of our class distribution estimation method.

\section{SAOT}
\label{sec:appsaot}
As mentioned in Section \ref{sec:saot}, the SAOT problem is formulated as:
\begin{equation}
    \begin{aligned}
        \min_{\bm{\pi},\bm{b}}{\langle \bm{\pi}, \bm{M} \rangle} + \epsilon_1 H(\bm{\pi}) + \epsilon_2 (\Psi(\bm{b}))^T\bm{1},\\
        s.t.\,\,\bm{\pi}\bm{1}=\bm{a},\bm{\pi}^T\bm{1}=\bm{b}, \bm{\pi}\geq0, \bm{b}^T\bm{1}=1.
    \end{aligned}
\end{equation}
where $\epsilon_1$ and $\epsilon_2$ are balance hyper-parameters, $\Psi(\bm{b})=-\log\bm{b} - \log(1-\bm{b})$ is the penalty function about $\bm{b}$.
We adopt the Lagrangian multiplier algorithm to optimize the problem:
\begin{equation}
\label{eq:lq}
    \begin{aligned}
        &\min_{\bm{\pi},\bm{b}}{\langle \bm{\pi}, \bm{M} \rangle} + \epsilon_1 H(\bm{\pi}) + \epsilon_2 (\Psi(\bm{b}))^T\bm{1}\\
        &-\bm{f}^T(\bm{\pi}\bm{1}-\bm{a})-\bm{g}^T(\bm{\pi}^T\bm{1}-\bm{b})-h(\bm{b}^T\bm{1}-1),
    \end{aligned}
\end{equation}
where $\bm{f}$, $\bm{g}$, and $h$ are all Lagrangian multipliers.
Taking the differentiation of Equation (\ref{eq:lq}) on the variable $\bm{\pi}$, we can obtain:
\begin{equation}
\label{eq:update_q}
    \begin{aligned}
        \pi_{ij} = \exp(\frac{f_i+g_j-M_{ij}}{\epsilon_1})>0.
    \end{aligned}
\end{equation}
We first fix $\bm{b}$, 
due to the fact that $\bm{\pi}\bm{1}=\bm{a}$ and $\bm{\pi}^T\bm{1}=\bm{b}$, we can get:
\begin{equation}
\label{eq:update_f}
    \begin{aligned}
        \exp{(\frac{f_i}{\epsilon_1})}=\frac{a_i}{\sum_j^C\exp(\frac{g_j-M_{ij}}{\epsilon_1})},
    \end{aligned}
\end{equation}
\begin{equation}
\label{eq:update_g}
    \begin{aligned}
        \exp{(\frac{g_j}{\epsilon_1})}=\frac{b_j}{\sum_i^N\exp(\frac{f_i-M_{ij}}{\epsilon_1})}.
    \end{aligned}
\end{equation}
Then we fix $\bm{f}$ and $\bm{g}$, and update $\bm{b}$ by:
\begin{equation}
\label{eq:lb}
    \begin{aligned}
        \min_{\bm{b}} \left[ {\epsilon_2(\Psi(\bm{b}))^T\bm{1}+\bm{g}^T\bm{b}-h(\bm{b}^T\bm{1}-1)}\right].
    \end{aligned}
\end{equation}
Taking the differentiation of Equation (\ref{eq:lb}) on the variable $\bm{b}$, we can obtain:
\begin{equation}
\label{eq:beq}
    \begin{aligned}
        (g_j-h)b_j^2-((g_j-h)+2\epsilon_2)b_j+\epsilon_2=0.
    \end{aligned}
\end{equation}
It is easy to get the discriminant of Equation (\ref{eq:beq}) $\Delta_j=(g_j-h)^2+4\epsilon_2^2>0$, 
\begin{equation}
    \begin{aligned}
        b_j(h)=\frac{(g_j-h+2\epsilon_2)\pm \sqrt{\Delta_j}}{2(g_j-h)}.
    \end{aligned}
\end{equation}
Note that,
\begin{equation}
\label{eq:update_b1}
    \begin{aligned}
        b_j(h)=\frac{((g_j-h)+2\epsilon_2)+\sqrt{\Delta_j}}{2(g_j-h)}\geq 1.
    \end{aligned}
\end{equation}
Thus, we choose the following $b_j(h)$:
\begin{equation}
\label{eq:update_b}
    \begin{aligned}
        b_j(h)=\frac{((g_j-h)+2\epsilon_2)- \sqrt{\Delta_j}}{2(g_j-h)}.
    \end{aligned}
\end{equation}
Taking Equation (\ref{eq:update_b}) back to the original constraint $\bm{b}^T\bm{1}=1$, the formula is defined as below:
\begin{equation}
\label{eq:fh}
    \begin{aligned}
        (\bm{b}(h))^T\bm{1}-1=0,
    \end{aligned}
\end{equation}
where $h$ is the root of Equation (\ref{eq:fh}), and we can use Newton's method to work out it. 
Specifically, we first define that $f(h)=(\bm{b}(h))^T\bm{1}-1$, then $h$ can be updated by:
\begin{equation}
\label{eq:update_h}
    \begin{aligned}
        h \gets h - \frac{f(h)}{f^{'}(h)},
    \end{aligned}
\end{equation}
where the iteration number is set to 10. 
Then we can obtain $\bm{b}$ by Equation (\ref{eq:update_b}).
In short, through iteratively updating Equation (\ref{eq:update_f}), (\ref{eq:update_g}), (\ref{eq:update_h}), and (\ref{eq:update_b}), we can obtain the transport matrix $\bm{\pi}$ on Equation (\ref{eq:update_q}). 
We show the iteration optimization scheme of SAOT in Algorithm~\ref{alg:saot}.

\begin{algorithm}[t]
\caption{The optimization scheme of SAOT }\label{alg:saot}
\textbf{Input:} The cost distance matrix: $\bm{M}$.\\
\textbf{Output:} The transport matrix: $\bm{\pi}$.\\
\textbf{Procedure}:
\begin{algorithmic}[1]
{
\STATE{Initialize $\bm{f}$ and $\bm{g}$ randomly;}
\STATE{Initialize $\bm{b}$ randomly and perform normalization so that $\bm{b}^T\bm{1}=1$;}
\STATE{Initialize $h=1$.}
\FOR{$i=1$ to $T$}
\STATE{Update $\bm{f}$ in Equation (\ref{eq:update_f});}
\STATE{Update $\bm{g}$ in Equation (\ref{eq:update_g});}
\STATE{Update $\bm{b}$ in Equation (\ref{eq:update_b}) with the constraint $\bm{b}^T\bm{1}=1$. 
}
\ENDFOR
\STATE{Calculate $\bm{\pi}$ in Equation (\ref{eq:update_q}).}
}
\end{algorithmic}
\end{algorithm}

\section{Experiment}
\subsection{Datasets}
\label{sec:appdata}
We conduct extensive experiments on eight popularly used real-world datasets.
The details of each dataset are as follows.

\textbf{AgNews} \cite{rakib2020enhancement} is a subset of AG's news corpus collected by \cite{zhang2015character} which consists of 8,000 news titles in 4 topic categories.
\textbf{StackOverflow} \cite{xu2017self} consists of  20,000 question titles associated with 20 different tags, which is randomly selected from the challenge data published in Kaggle.com\footnote{https://www.kaggle.com/c/predict-closed-questions-on-stack-overflow/}.
\textbf{Biomedical} \cite{xu2017self} is composed of 20,000 paper titles from 20 different topics and it is selected from the challenge data published in BioASQ's official website\footnote{http://participants-area.bioasq.org/}.
\textbf{SearchSnippets} \cite{phan2008learning} contains 12,340 snippets from 8 different classes, which is selected from the results of web search transaction.
\textbf{GoogleNews} \cite{yin2016model} consists of the titles and snippets of 11,109 news articles about 152 events \cite{yin2014dirichlet} which is divided into three datasets: the full dataset is \textbf{GoogleNews-TS}, while \textbf{GoogleNews-T} only contains titles and \textbf{GoogleNews-S} only has snippets.
\textbf{Tweet} \cite{yin2016model} consists of 2,472 tweets related to 89 queries and the original data is from the 2011 and 2012 microblog track at the Text REtrieval Conference\footnote{https://trec.nist.gov/data/microblog.html}.
The detailed statistics of these datasets are shown in Table \ref{ta:dataset}.

\begin{table}
\centering
	\begin{tabular}{c | c c c c c}
	\hline
		Dataset & C & N & A & R \\
		\hline
		AgNews & 4 & 8,000 & 23 & 1\\
		StackOverflow & 20 & 20,000 & 8 & 1\\
		Biomedical & 20 & 20,000 & 13 & 1\\
		SearchSnippets & 8 & 12,340 & 18 & 7\\
		GoogleNews-TS & 152 & 11,109 & 28 & 143\\
		GoogleNews-T & 152 & 11,108 & 6 & 143\\
		GoogleNews-S & 152 & 11,108 & 22 & 143\\
		Tweet & 89 & 2,472 & 9 & 249\\
		\hline
	\end{tabular}
	\caption{The statistics of the datasets. C means the number of classes, N means the dataset size, A is the average number of words per instance and R is the ratio of the largest cluster size to the smallest one. }
	\label{ta:dataset}
\end{table}

\subsection{Experiment Settings}
\label{sec:appimple}
We choose distilbert-base-nli-stsb-mean-tokens in Sentence Transformer library \cite{reimers2019sentence} to encode the text, and the maximum input length is set to $32$.
The learning rate is set to $5\times10^{-6}$ for optimizing the encoding network, and $5\times10^{-4}$ for optimizing both the projecting network and clustering network.
The dimensions of the text representations and the projected representations are set to $D_1=768$ and $D_2=128$, respectively.
The batch size is set to $N=200$.
The temperature parameter is set to $\tau=1$.
The threshold $\delta$ is set to $0.01$.
The datasets specific tuning is avoided as much as possible.
For \textbf{BOW} and \textbf{TF-IDF}, we achieved the code with scikit-learn \cite{scikit-learn}.
For all the other baselines, i.e., \textbf{STC$^2$-LPI}\footnote{https://github.com/jacoxu/STC2},
\textbf{Self-Train}
\footnote{https://github.com/hadifar/stc$\_$clustering},
\textbf{K-means$\_$IC}\footnote{https://github.com/rashadulrakib/short-text-clustering-enhancement}, and
\textbf{SCCL}\footnote{https://github.com/amazon-science/sccl} (MIT-0 license), we used their released code.
Besides, we substitute the accuracy evaluation code of \textbf{K-means$\_$IC} with the evaluation method described in our paper.

In addition, as \textbf{STC$^2$-LPI} and \textbf{Self-Train} use the word embeddings pre-trained with in-domain corpus, and there are only three datasets' pre-trained word embeddings provided, therefore we do not report the results of other five datasets for them.

\subsection{Evaluation Metrics}
\label{sec:appeval}
We report two widely used evaluation metrics of text clustering, i.e., accuracy (ACC) and normalized mutual information (NMI), following \cite{xu2017self, hadifar2019self, zhang2021supporting}.
Accuracy is defined as:
\begin{equation}
    \begin{aligned}
        ACC=\frac{\sum_{i=1}^N{\mathbbm{1}_{y_i=map(\hat{y}_i)}}}{N},
    \end{aligned}
\end{equation}
where $y_i$ and $\hat{y}_i$ are the ground truth label and the predicted label for a given text $x_i$ respectively, $map()$ maps each predicted label to the corresponding target label by Hungarian algorithm\cite{papadimitriou1998combinatorial}. 
Normalized mutual information is defined as:
\begin{equation}
    \begin{aligned}
        NMI(Y, \hat{Y})=\frac{I(Y, \hat{Y})}{\sqrt{H(Y)H(\hat{Y})}},
    \end{aligned}
\end{equation}
where $Y$ and $\hat{Y}$ are the ground truth labels and the predicted labels respectively, $I()$ is the mutual information, and $H()$ is the entropy.

\begin{figure}[t]
  \centering
  \includegraphics[width=5cm]{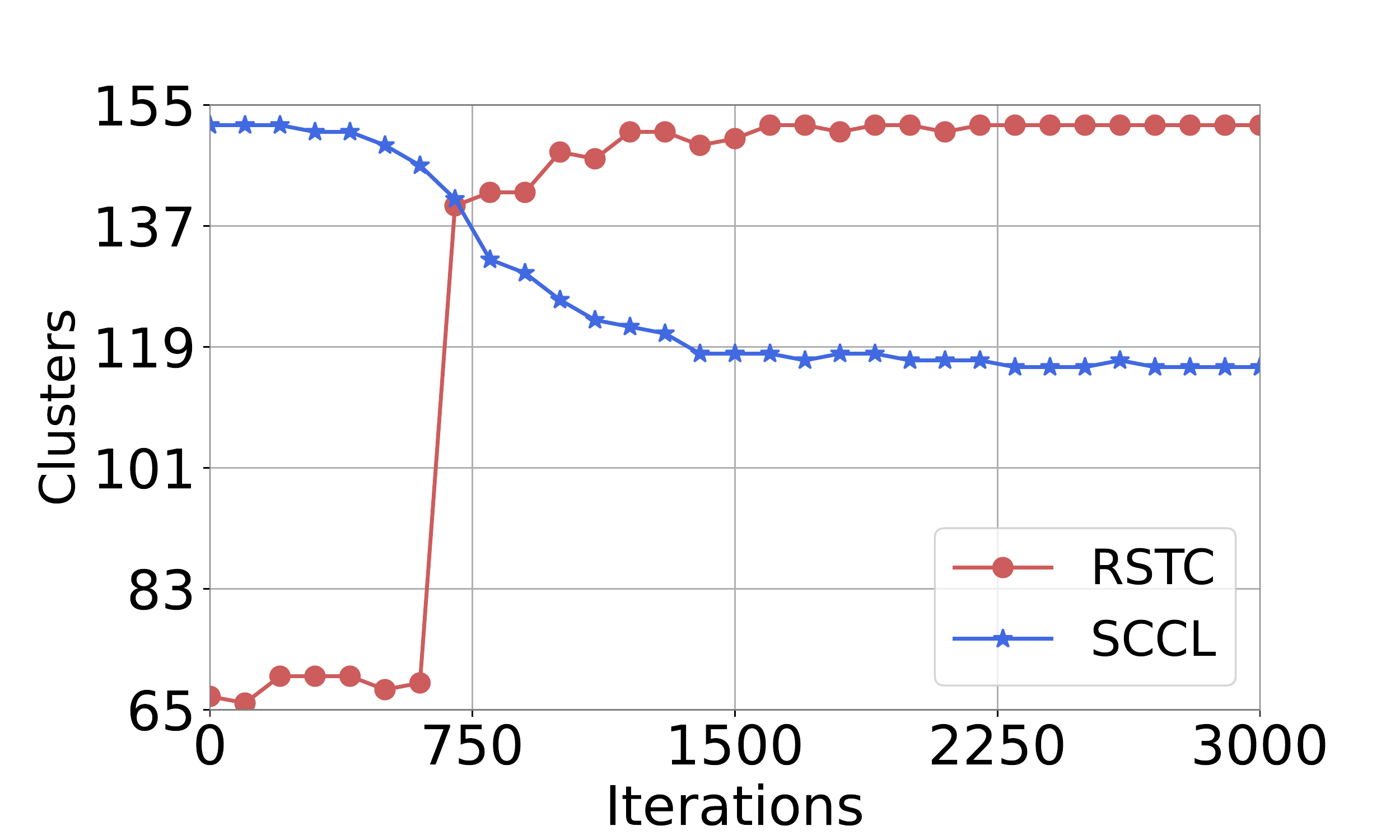}
  \caption{The number of predicted clusters at different iterations on GoogleNews-T.
  The steep rise of the \textit{clusters} for \modelname~is because we first adopt 600 steps of $\mathcal{L}_I$ training and then adopt the training of both $\mathcal{L}_C$ and $\mathcal{L}_I$, for a pre-training representations purpose.
  }
  \label{fig:scclresult}
\end{figure}

\subsection{Visualization}
\label{sec:appvisual}
To better show the clustering degeneracy problem, we visualize how the number of predicted clusters (we call it \textit{clusters} later) are changing over iterations on \textbf{SCCL} and \modelname.
The results are shown in Fig.\ref{fig:scclresult}.
From it, we verify that \textbf{SCCL} has relatively serious clustering degeneracy problem while \modelname~solves it to some extent.
Specifically, the \textit{clusters} of \textbf{SCCL} is much less than the real category number.
Moreover, the degeneracy has a negative effect on the final k-means clustering performance because it makes the representations getting worse.
Whereas the \textit{clusters} of \modelname~almost convergent to real category number, which assures the high accuracy of \modelname.
The visualization results illustrate the validity of our model.

\subsection{Computational Budget}
The number of parameters in our model is 68M. Our training for each dataset takes about 10-30 minutes, using a GeForce RTX 3090 GPU.
\end{document}